\documentclass[journal]{IEEEtran}
\usepackage{times}
\usepackage{epsfig}
\usepackage{graphicx}
\usepackage{subfigure}
\usepackage{amsmath}
\usepackage{amssymb}
\usepackage{color}
\usepackage[pagebackref=false,breaklinks=true,letterpaper=true,colorlinks,bookmarks=false]{hyperref}
\usepackage{cite}
\usepackage{caption}
\usepackage{makecell}
\usepackage{array}
\usepackage{float}
\usepackage{bm}
\usepackage{multirow}
\usepackage{indentfirst}
\usepackage{url}
\usepackage{balance}
\begin{document}
\title{Remote Sensing Image Scene Classification Meets Deep Learning: Challenges, Methods, Benchmarks, and Opportunities}

\author{Gong Cheng, Xingxing Xie, Junwei Han,~\IEEEmembership{Senior Member, IEEE}, Lei Guo, Gui-Song Xia,~\IEEEmembership{Senior Member, IEEE}
	
\thanks{This work was supported in part by the Science, Technology and Innovation Commission of Shenzhen Municipality under Grant JCYJ20180306171131643, in part by the National Science Foundation of China under Grants 61772425 and 61773315, and in part by the Fundamental Research Funds for the Central Universities under Grant 3102019AX09 (J. Han is the corresponding author).}
\thanks{G. Cheng and X. Xie are with the Research \& Development Institute of Northwestern Polytechnical University in Shenzhen, Shenzhen 518057, China, and also with the School of Automation, Northwestern Polytechnical University, Xi'an 710129, China.}
\thanks{J. Han and L. Guo are with the School of Automation, Northwestern Polytechnical University, Xi'an 710129, China.}
\thanks{G.-S. Xia is with the School of Computer Science, Wuhan University, Wuhan 430072, China.}
\thanks{Manuscript received xxx xx, xxx.}}

\markboth{Journal of \LaTeX\ Class Files,~Vol.~x, No.~x, xxxx~2019}%
{Shell \MakeLowercase{\textit{et al.}}: Bare Demo of IEEEtran.cls for IEEE Journals}

\maketitle
\begin{abstract}
Remote sensing image scene classification, which aims at labeling remote sensing images with a set of semantic categories based on their contents, has broad applications in a range of fields. Propelled by the powerful feature learning capabilities of deep neural networks, remote sensing image scene classification driven by deep learning has drawn remarkable attention and achieved significant breakthroughs. However, to the best of our knowledge, a comprehensive review of recent achievements regarding deep learning for scene classification of remote sensing images is still lacking. Considering the rapid evolution of this field, this paper provides a systematic survey of deep learning methods for remote sensing image scene classification by covering more than 160 papers. To be specific, we discuss the main challenges of remote sensing image scene classification and survey (1) Autoencoder-based remote sensing image scene classification methods, (2) Convolutional Neural Network-based remote sensing image scene classification methods, and (3) Generative Adversarial Network-based remote sensing image scene classification methods. In addition, we introduce the benchmarks used for remote sensing image scene classification and summarize the performance of more than two dozen of representative algorithms on three commonly-used benchmark data sets. Finally, we discuss the promising opportunities for further research.
\end{abstract}



%
\IEEEpeerreviewmaketitle

\section{Introduction}
\IEEEPARstart{R}{emote} sensing images, a valuable data source for earth observation, can help us to measure and observe detailed structures on the Earth's surface. Thanks to the advances of earth observation technology \cite{hu2013exploring,gomez2015multimodal}, the volume of remote sensing images is drastically growing. This has given particular urgency to the quest for how to make full use of ever-increasing remote sensing images for intelligent earth observation \cite{gamba2012human,li2017earth}. Hence, it is extremely important to understand huge and complex remote sensing images.
\begin{figure}[th]
	\centering
	\includegraphics[width=\linewidth]{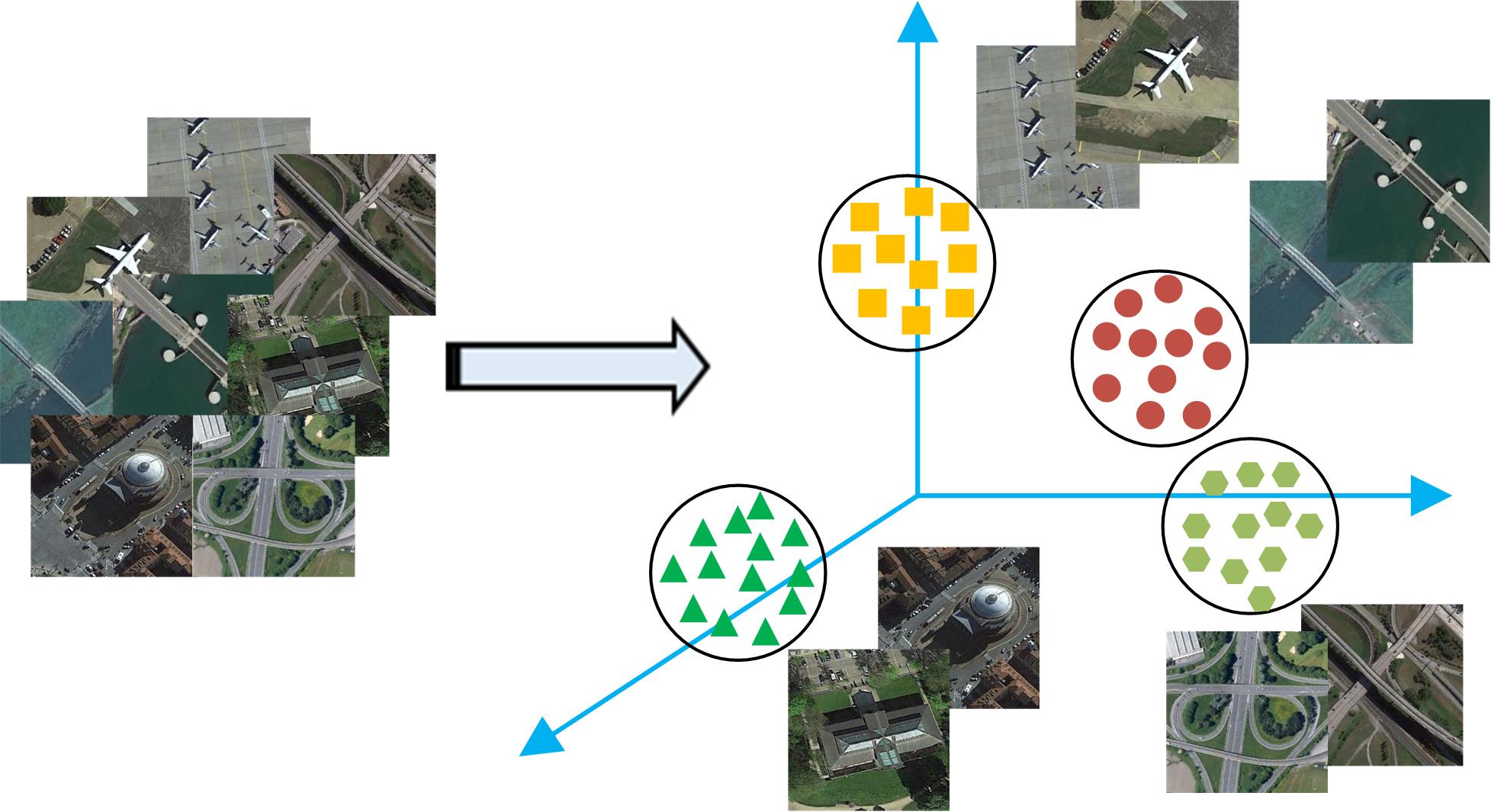}\\
	\caption{Illustration of remote sensing  image  scene classification, which aims at labeling each remote sensing image patch with a semantic class based on its content.}
	\label{fig1}
\end{figure}
As a key and challenging problem for effectively interpreting remote sensing imagery, scene classification of remote sensing images has been an active research area. Remote sensing image scene classification is to correctly label given remote sensing images with predefined semantic categories, as shown in Fig. \ref{fig1}. For the last few decades, extensive researches on  remote sensing image scene classification have been undertaken driven by its real-world applications, such as urban planning \cite{longbotham2011very,tayyebi2011urban}, natural hazards detection \cite{martha2011segment,ChengAutomatic,lv2018landslide}, environment monitoring\cite{huang2017multi,zhang2018monitoring,ghazouani2019a}, vegetation mapping \cite{li2013object,mishra2014mapping}, and geospatial object detection \cite{cheng2016learning,li2018deep,cheng2018learning,cheng2016survey,li2020object,li2017rotation,cheng2016rifd,cheng112015learning}.

\begin{figure*}[th]
	\centering
	\includegraphics[width=\linewidth]{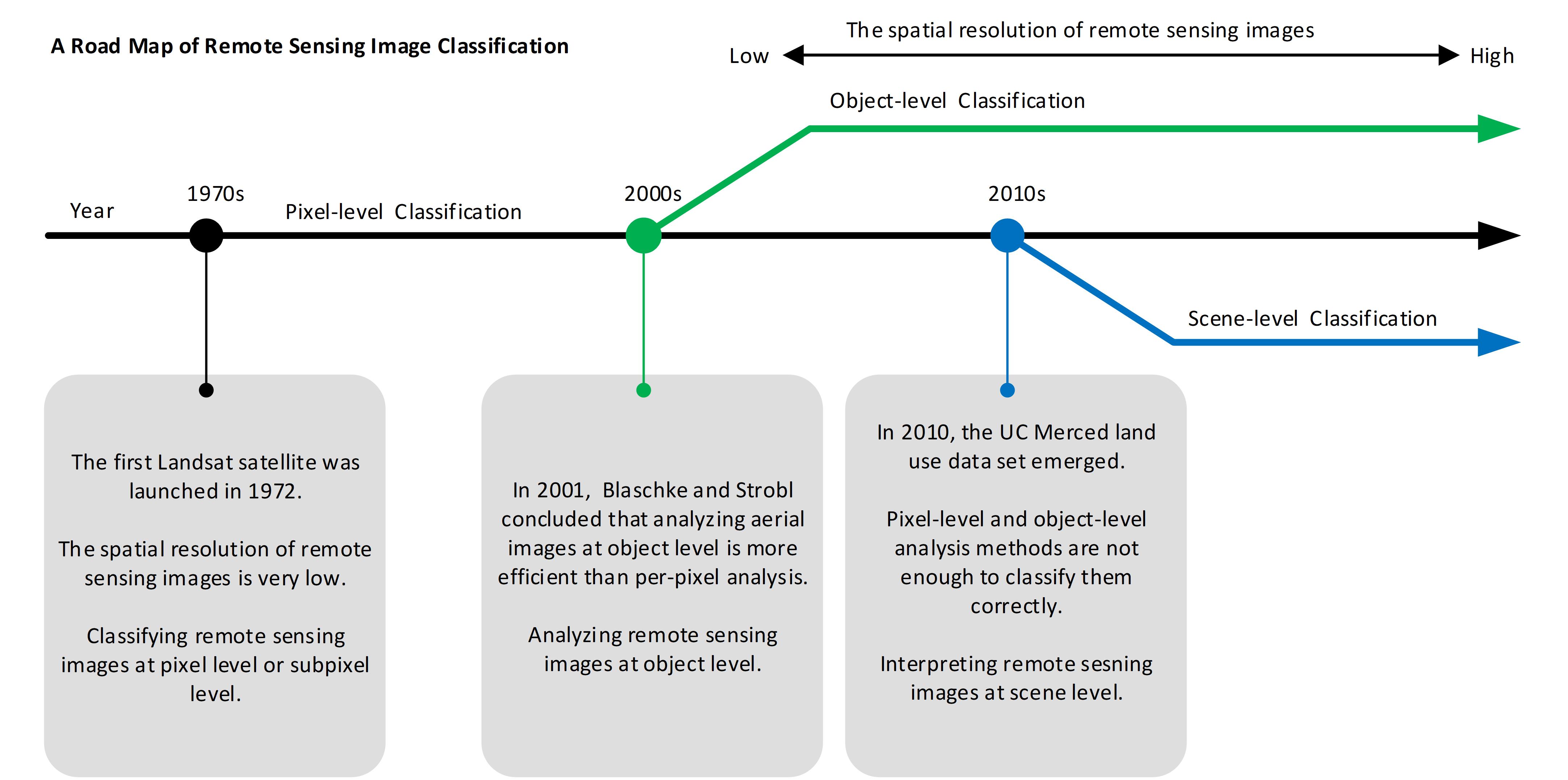}\\
	\caption{ A road map of remote sensing image classification. With the improvement of spatial resolution of remote sensing images, remote sensing image classification gradually formed three parallel classification branches at different levels: pixel-level, object-level, and scene-level classification. Here, it is worth mentioning that we use ``remote sensing image classification'' as a general concept.}
	\label{fig2}
\end{figure*}

\begin{figure}[th]
	\centering
	\includegraphics[width=\linewidth]{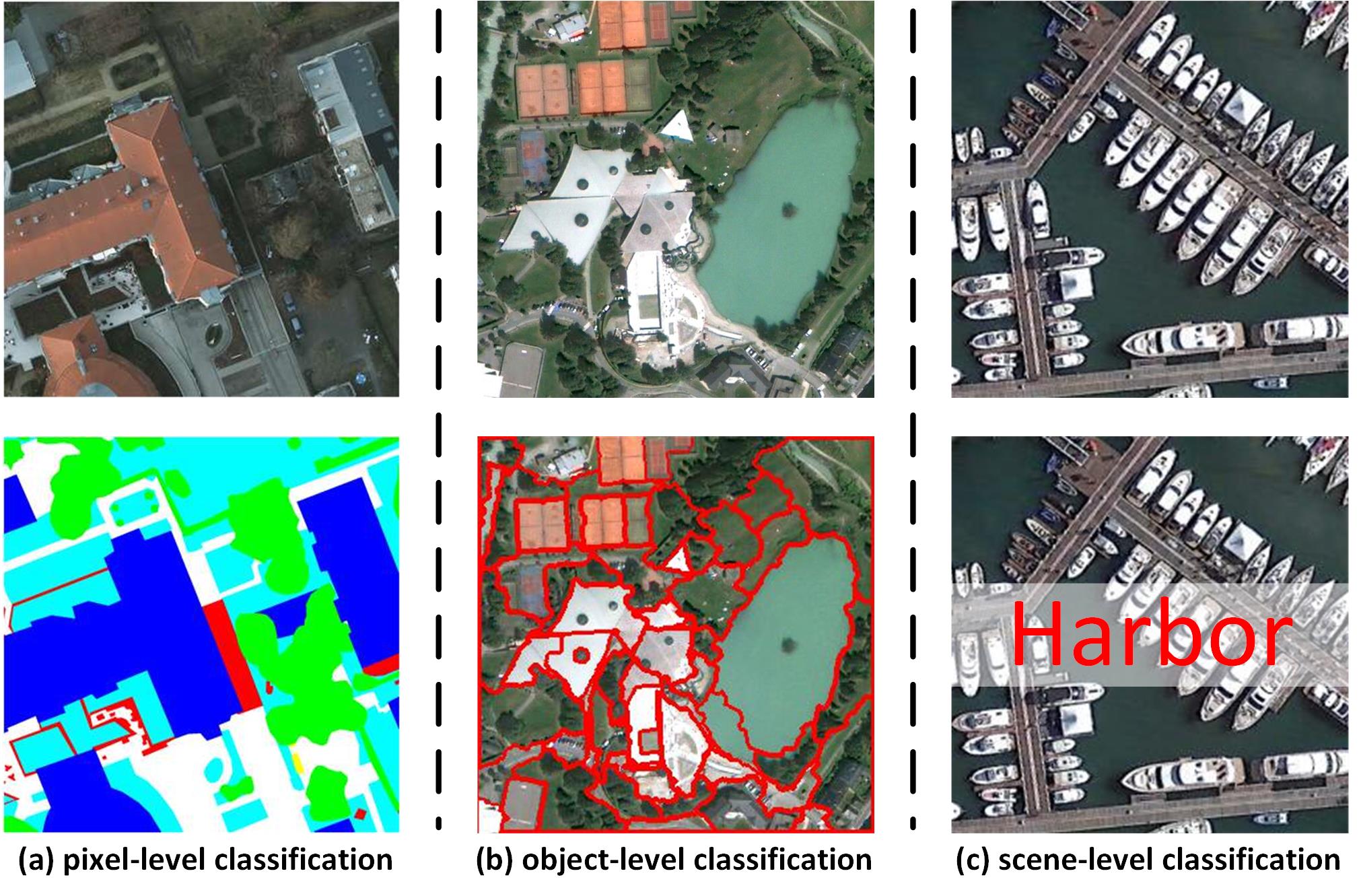}\\
	\caption{ Three levels of remote sensing image classification: (a) pixel-level remote sensing image classification focuses on labeling each pixel with a class; (b) object-level remote sensing image classification aims at recognizing objects in remote sensing images; (c) scene-level remote sensing image classification seeks to classify each given remote sensing image patch into a semantic class. This survey focuses on scene-level remote sensing image classification.}
	\label{fig3}
\end{figure}

With the improvement of spatial resolution of remote sensing images, remote sensing image classification gradually formed three parallel classification branches at different levels: pixel-level, object-level, and scene-level classification, as shown in Fig. \ref{fig2} and Fig. \ref{fig3}. Here, it is worth mentioning that we use the term of ``remote sensing image classification'' as a general concept, which includes pixel-level, object-level, and scene-level classification of remote sensing images. To be specific, in the early literatures, researchers mainly focused on classifying remote sensing images at pixel level or subpixel level\cite{ji1999effectiveness,tuia2009active,tuia2011survey}, through labeling each pixel in the remote sensing images with a semantic class, because the spatial resolution of remote sensing images is very low--the size of a pixel is similar to the sizes of the objects of interest\cite{janssen1992knowledge}. To date, pixel-level remote sensing image classification (sometimes also called semantic segmentation, as shown in Fig. \ref{fig3} (a)) is still an active research topic in the areas of multispectral and hyperspectral remote sensing image analysis \cite{ghamisi2017advanced,he2017recent,li2019deep,cheng2018exploring,zhou2019learning}.

Due to the advance of remote sensing imaging, the spatial resolution of remote sensing images is increasingly finer than common objects of interest, such that single pixels lose their semantic meanings. In such case, it is not feasible to recognize scene images at the pixel level solely and so per-pixel analysis began to be viewed with increasing dissatisfaction. In 2001, Blaschke and Strobl\cite{blaschke2001s} questioned the dominance of per-pixel research paradigm and concluded that analyzing remote sensing images at the object level is more efficient than per-pixel analysis. They suggested that researchers should pay attention to object-level analysis, which aims at recognizing objects in remote sensing images, as shown in Fig. \ref{fig3} (b), where the term ``object'' refers to meaningful semantic entities or scene units. Subsequently, a series of approaches to analyze remote remote sensing images at object level has dominated remote sensing image analysis for the last two decades\cite{blaschke2003object,yan2006comparison,blaschke2010object,blaschke2008object}. Amazing achievements of certain specific land use identification tasks have been accomplished by pixel-level and object-level classification algorithms.

However, remote sensing images may contain different and distinct object classes because of the increasing resolutions of remote sensing images. Pixel-level and object-level methods may not be sufficient to always classify them correctly. Under the circumstances, it is of considerable interest to understand the global contents and meanings of remote sensing images. A new paradigm of scene-level analysis of remote sensing images has been recently suggested. Scene-level remote sensing image classification, namely remote sensing image scene classification, seeks to classify each given remote sensing image patch (e.g., 256$\times$256) into a semantic class, as illustrated in Fig. \ref{fig3} (c). Here the item ``scene'' represents an image patch cropped from a large-scale remote sensing image that contains clear semantic information on the earth surface\cite{cheng2015effective,hu2015transferring}.

It is a significant step to be able to represent visual data with discriminative features in almost all tasks of computer vision. The remote sensing domain is no exception. During the previous decade, extensive efforts have been devoted to developing discriminative visual features. A majority of early remote sensing image scene classification methods relied on human-engineering descriptors, e.g., Scale-Invariant Feature Transformation (SIFT)\cite{lowe2004distinctive}, Texture Descriptors (TD)\cite{haralick1973textural,jain1997object,ojala2002multiresolution}, Color Histogram (CH)\cite{swain1991color}, Histogram of Oriented Gradients (HOG)\cite{dalal2005histograms}, and GIST\cite{oliva2001modeling}. Owing to their characteristic of being able to represent an entire image with features, it is feasible to directly apply CH, GIST and TD to remote sensing image scene classification. However, SIFT and HOG cannot represent an entire image directly because of their local characteristic. To make handcrafted local descriptors represent an entire scene image, these local descriptors are encoded by certain encoding methods (e.g., the Improved Fisher Kernel (IFK)\cite{perronnin2010improving}, Vector of Locally Aggregated Descriptors (VLADs)\cite{jegou2011aggregating}, Spatial Pyramid Matching (SPM)\cite{lazebnik2006beyond}, and the popular Bag-Of-Visual-Words (BoVW)\cite{yang2010bag}). Thanks to the simplicity and efficiency of these feature encoding methods, they have been broadly applied to the field of remote sensing image scene classification\cite{yang2011spatial,shao2013hierarchical,negrel2014evaluation,zhao2014land,zhang2013high,zhu2016bag}, whereas the representation capability of handcrafted features is limited.
\\ \indent
In this case, unsupervised learning, such as k-means clustering, Principal Component Analysis (PCA)\cite{wold1987principal}, and sparse coding\cite{olshausen1997sparse}, which automatically learns features from unlabeled images, become an appealing alternative to human-engineering features. A considerable amount of unsupervised learning-based scene classification methods have emerged\cite{cheriyadat2013unsupervised,mekhalfi2015land,risojevic2016unsupervised,sheng2012high,zheng2012automatic,zhong2015scene,lu2017remote,fan2017unsupervised,romero2015unsupervised}, and made substantial progress for scene classification. Nevertheless, these unsupervised learning approaches cannot make full use of data class information. \\ \indent Fortunately, due to the advances in deep learning theory and the increased availability of remote sensing data and parallel computing resources, deep learning-based algorithms have increasingly prevailed the area of remote sensing image scene classification. In 2006, Hinton and Salakhutdinov\cite{hinton2006reducing} created an approach to initialize the weights for training multilayer neural networks, which builds a solid foundation for the development of deep learning later. During the period 2006 to 2012, simple deep learning models have been developed (e.g., deep belief nets\cite{hinton2006fast}, autoencoder\cite{hinton2006reducing}, and stacked autoencoder\cite{vincent2010stacked}).

The feature description capabilities of these simple deep learning models have been demonstrated in many fields, involving remote sensing image scene classification. Since the AlexNet, a deep Convolutional Neural Network (CNN) designed by Krizhevskey et al. \cite{krizhevsky2012imagenet} in 2012, obtained the best results in the Large-Scale Visual Recognition Challenge (LSVRC) \cite{deng2009imagenet},
a great many advanced deep CNNs have come forth and broken a number of records in many fields. In the wake of these successes, CNN-based methods have emerged in remote sensing image scene classification\cite{minetto2019hydra,cheng2018deep,wang2018scene} and achieved advanced classification accuracy.
\begin{figure}[th]
	\centering
	\includegraphics[width=\linewidth]{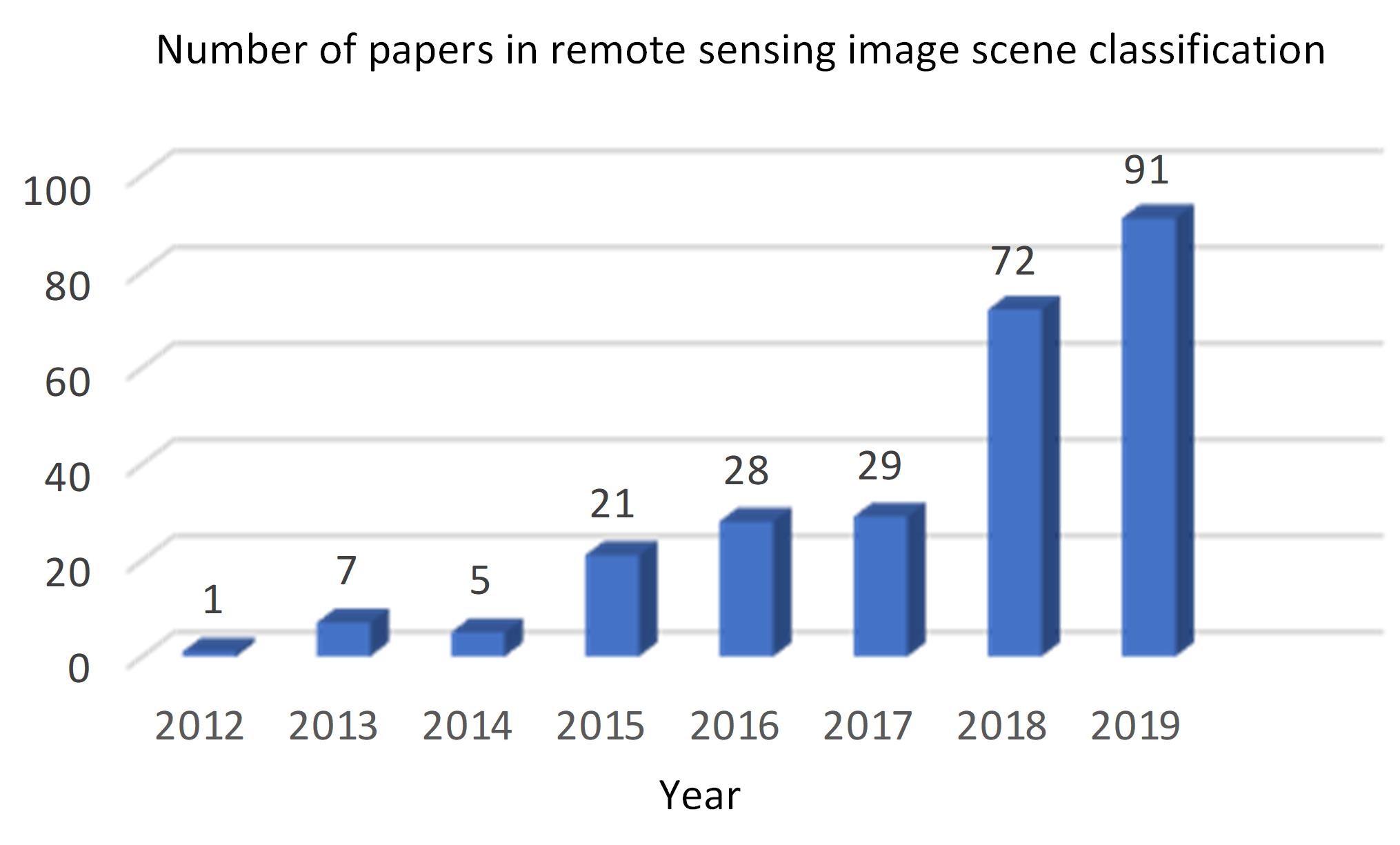}\\
	\caption{The number of publications in remote sensing image scene classification from 2012 to 2019. Data from google scholar advanced search: allintitle: (``remote sensing'' or ``aerial'' or ``satellite'' or ``land use'') and ``scene classification''.}
	\label{fig4}
\end{figure}

\begin{table*}[th]
	\centering
1	\caption{Summarization of a number of surveys of remote sensing image analysis.}\label{table1}
	\resizebox{0.97\textwidth}{!}{
		\renewcommand\arraystretch{1.6}
		\begin{tabular}{|p{0.5cm}<{\centering}|p{6.1cm}<{\centering}|p{1.0cm}<{\centering}|p{2.5cm}<{\centering}|p{6.1cm}<{\centering}|}
			\hline
			No. & Survey Title & Year & Publication & Content\\
			\hline
			1 & \multicolumn{1}{m{6cm}<{\centering}|}{A survey of active learning algorithms for supervised remote sensing image classification \cite{tuia2011survey}}
			& 2011 & IEEE JSTSP
			&\multicolumn{1}{m{6cm}<{\centering}|}{Surveying and testing the main families of active learning methods}\\
			
			\hline
			2 &\multicolumn{1}{m{6cm}<{\centering}|}{A review of remote sensing image classification techniques: the role of spatio-contextual information \cite{li2014review}}
			& 2014 & EuJRS
			& \multicolumn{1}{m{6cm}<{\centering}|}{Review of pixel-wise, subpixel-wise and object-based methods for remote sensing image classification and exploring the contribution of spatio-contextual information to scene classification}  \\
			
			\hline
			3 &\multicolumn{1}{m{6cm}<{\centering}|}{Multimodal classification of remote sensing images: a review and future directions \cite{gomez2015multimodal}}
			& 2015 &\multicolumn{1}{m{2.5cm}<{\centering}|}{Proceedings of the IEEE}
			&\multicolumn{1}{m{6cm}<{\centering}|}{Offering a taxonomical view of the field of multimodal remote sensing image classification}  \\
			
			\hline
			4 &\multicolumn{1}{m{6cm}<{\centering}|}{Deep learning for remote sensing data: A technical tutorial on the state of the art \cite{zhang2016deep}}
			& 2016 & IEEE GRSM
			&\multicolumn{1}{m{6cm}<{\centering}|}{Reviewing deep learning-based remote sensing data analysis techniques before 2016}  \\
			
			\hline
			5 &\multicolumn{1}{m{6cm}<{\centering}|}{Deep learning in remote sensing: A comprehensive review and list of resources \cite{zhu2017deep}}
			& 2017 & IEEE GRSM
			& \multicolumn{1}{m{6cm}<{\centering}|}{Reviewing the progress of deep learning-based remote sensing data analysis before 2017} \\
			
			\hline
			6 &\multicolumn{1}{m{6cm}<{\centering}|}{Advanced spectral classifiers for hyperspectral images: A review \cite{ghamisi2017advanced} }
			& 2017 & IEEE GRSM
			&\multicolumn{1}{m{6cm}<{\centering}|}{ Review and comparison of different supervised hyperspectral classification methods}  \\
			
			\hline
			7 &\multicolumn{1}{m{6.5cm}<{\centering}|}{Remote sensing image classification: a survey of support-vector-machine-based advanced techniques \cite{maulik2017remote}}
			& 2017 & IEEE GRSM
			&\multicolumn{1}{m{6cm}<{\centering}|}{Review of remote sensing image classification based on SVM}  \\
			
			\hline
			
			8 &\multicolumn{1}{m{6cm}<{\centering}|}{AID: a benchmark data set for performance evaluation of remote sensing image scene classification \cite{xia2017aid}}
			&  2017  & IEEE TGRS
			&\multicolumn{1}{m{6cm}<{\centering}|}{Review of aerial image scene classification methods before 2017 and proposing the AID data set}  \\
			
			\hline
			9 &\multicolumn{1}{m{6cm}<{\centering}|}{Remote sensing image scene classification: benchmark and state of the art \cite{cheng2017remote}}
			& 2017 &\multicolumn{1}{m{2.5cm}<{\centering}|}{Proceedings of the IEEE}
			&\multicolumn{1}{m{6cm}<{\centering}|}{Reviewing the progress of scene classification of remote sensing images before 2017 and proposing the NWPU-RESISC45 data set}  \\
			
			\hline
			10 &\multicolumn{1}{m{6cm}<{\centering}|}{Recent advances on spectral--spatial hyperspectral image classification: An overview and new guidelines \cite{he2017recent}}
			&  2017 &  IEEE TGRS
			&\multicolumn{1}{m{6cm}<{\centering}|}{ Survey of the progress in the classification of spectral--spatial hyperspectral images}  \\
			
			\hline
			11 &\multicolumn{1}{m{6cm}<{\centering}|}{ Deep learning for hyperspectral image classification: An overview \cite{li2019deep}}
			& 2019 & IEEE TGRS
			&\multicolumn{1}{m{6cm}<{\centering}|}{Review of hyperspectral image classification based on deep learning}  \\
			
			\hline
			12 &\multicolumn{1}{m{6cm}<{\centering}|}{Deep learning in remote sensing applications: A meta-analysis and review \cite{ma2019deep}}
			& 2019  & ISPRS JPRS
			&\multicolumn{1}{m{6cm}<{\centering}|}{Providing a review of the applications of deep learning in remote sensing image analysis} \\
			
			\hline
			\textbf{13} &\multicolumn{1}{m{6cm}<{\centering}|}{\textbf{Remote sensing image scene classification meets deep learning: challenges, methods, benchmarks, and opportunities}}
			& \textbf{2020} & \textbf{IEEE JSTARS}
			&\multicolumn{1}{m{6cm}<{\centering}|}{\textbf{A systematic review of recent advances in remote sensing image scene classification driven by deep learning}}  \\
			\hline
	\end{tabular}}
\end{table*}
Nevertheless, CNN-based methods generally demand massive annotated training data, which greatly limits their application scenarios. More recently, Generative Adversarial Networks (GANs)\cite{goodfellow2014generative}, a promising unsupervised learning method, have achieved significant success in many applications. To remedy the above-mentioned limitations, GANs have been employed by some researchers on the field of remote sensing image scene classification\cite{duan2018gan,lin2017marta}. \\ \indent Currently, driven by deep learning, a great number of methods of remote sensing image scene classification have sprung up (see Fig. \ref{fig4}). The number of papers in remote sensing image scene classification dramatically increased after 2014 and 2017 respectively. There are two reasons for the increase. On one hand, around 2014, deep learning techniques began to be applied to  remote sensing data analysis. On the other hand, in 2017, large-scale remote sensing image scene classification benchmarks appeared, which have greatly facilitated the development of deep learning-based remote sensing image scene classification.

In the past several years, numerous reviews of remote sensing image classification methods have been published, which are summarized in Table \ref{table1}. For example, Tuia et al. \cite{tuia2011survey} surveyed, tested and compared three active learning-based remote sensing image scene classification methods: committee, large margin, and posterior probability. G\'{o}Chova et al. \cite{gomez2015multimodal} surveyed multimodal remote sensing image classification and summarized the leading algorithms for this field. In\cite{maulik2017remote}, Maulik et al. conducted a review of remote sensing image scene classification algorithms based on support vector machine (SVM). Li et al. \cite{li2014review} surveyed the pixel-level, subpixel-level and object-based methods of image classification and emphasized the contribution of spatio-contextual information to remote sensing image scene classification.

As an alternative ways to extract robust, abstract and high-level features from images, deep learning models have made amazing progress on a broad range of tasks in processing image, video, speech and audio. After this, a number of deep learning-based scene classification algorithms were proposed, such as CNN-based methods and GAN-based methods. A number of reviews of scene classification approaches have been published. Penatti et al. \cite{penatti2015deep} assessed the generalization ability of pre-trained CNNs in classification of remote sensing images. In \cite{hu2015transferring}, Hu et al. surveyed how to apply the CNNs that trained on the ImageNet data set to remote sensing image scene classification. Zhu et al. \cite{zhu2017deep} presented a tutorial about deep learning-based remote sensing data analysis. In order to make full use of pre-trained CNNs, Nogueira et al. \cite{nogueira2017towards} analyzed the performance of CNNs for remote sensing image scene classification with different learning strategies: full training, fine tuning, and using CNNs as feature extractors. In \cite{zhang2016deep}, Zhang et al. reviewed the recent deep learning-based remote sensing data analysis. Considering the number of scene categories and the accuracy saturation of the existing scene classification data sets, Cheng et al. \cite{cheng2017remote} released a large-scale scene classification benchmark, named NWPU-RESISC45, and provided a survey of recent advance in remote sensing image scene classification before 2017. In \cite{xia2017aid}, Xia et al. proposed a novel benchmark, called AID, for aerial image classification and reviewed the existing methods of scene classification before 2017. Ma et al. \cite{ma2019deep} provided a review of the applications of deep learning in remote sensing image analysis. In addition, there have been several hyperspectral image classification surveys \cite{ghamisi2017advanced,he2017recent,li2019deep}.

However, a thorough survey of deep learning for scene classification is still lacking. This motivates us to deeply analyze the main challenges faced for remote sensing image scene classification, systematically review those deep learning-based scene classification approaches, most of which are published during the last five years, introduce the mainstream scene classification benchmarks, and discuss several promising future directions of scene classification.

The remainder of this paper is organized as follows. Section II discusses the current main challenges of remote sensing image scene classification. A brief review of deep learning models and a comprehensive survey of deep learning-based scene classification methods are provided in Section III. The scene classification data sets are introduced in Section IV. In Section V the comparison and discussion of the performance of deep learning-based scene classification methods on three widely used scene classification benchmarks are given. In Section VI, we discuss the promising future directions of scene classification. Finally, we conclude this paper in Section VII.

\section{Main Challenges of Remote Sensing Image Scene Classification}
\begin{figure*}[!th]
	\centering
	\includegraphics[width=\linewidth]{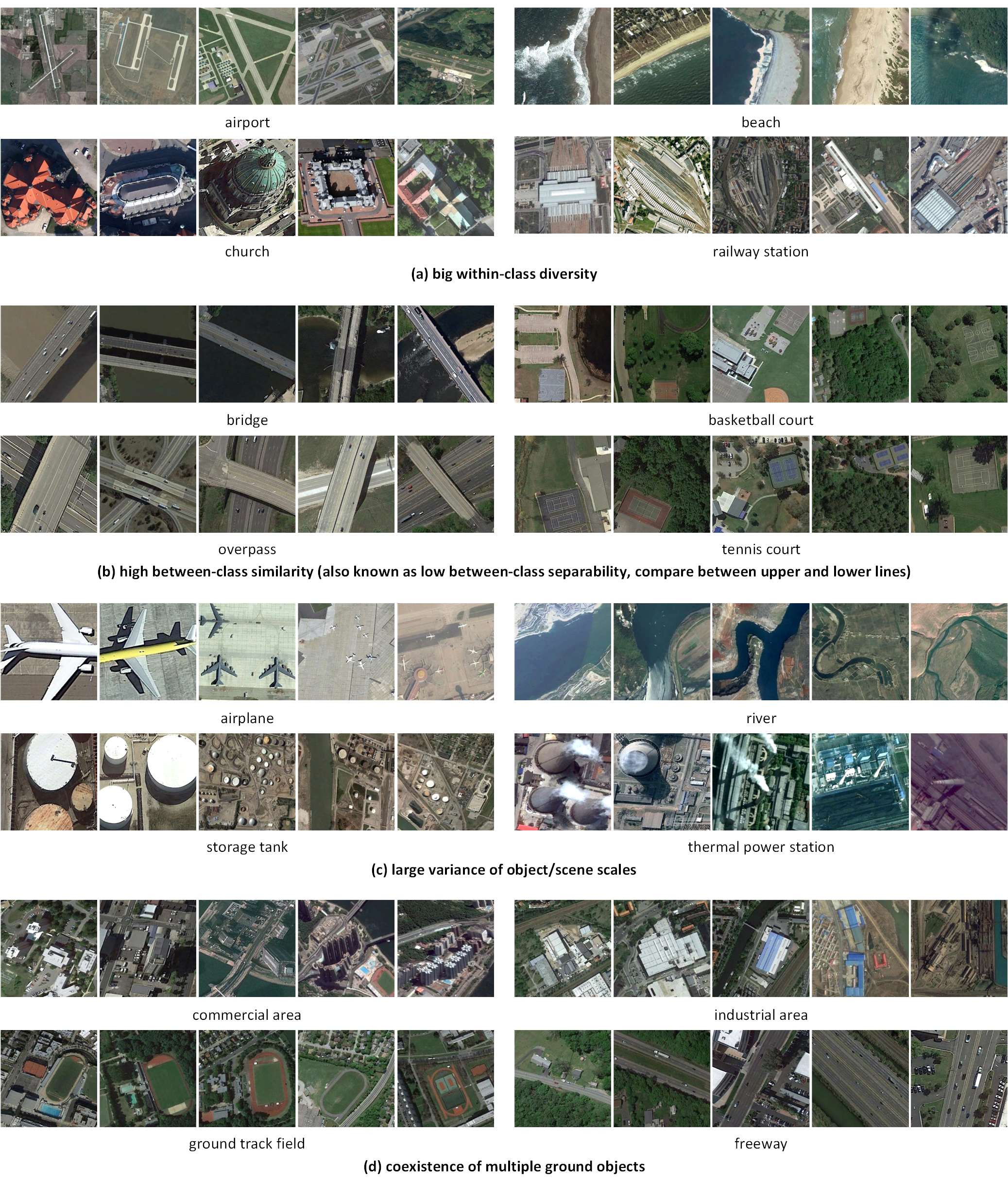}\\
	\caption{Challenges of remote sensing image scene classification, which include (a) big within-class diversity, (b) high between-class similarity (also known as low between-class separability), (c) large variance of object/scene scales, and (d) coexistence of multiple ground objects. These images are from the NWPU-RESISC45 data set \cite{cheng2017remote}.}
	\label{fig5}
\end{figure*}

The ideal goal of scene classification of remote sensing images is to correctly label the given remote sensing images with their corresponding semantic classes according to their contents, for example, categorizing a remote sensing image from urban into residential, commercial, or industrial area. Generally speaking, a remote sensing image contains a variety of ground objects. For instance, roads, trees, and buildings may be included in an industrial scene. Different from object-oriented classification, scene classification is a considerably challenging problem because of the variance and complex spatial distributions of ground objects existing in the scenes. Historically, extensive studies of remote sensing image scene classification have been made. However, there has not yet been an algorithm that can achieve the goal of classifying remote sensing image scenes with satisfactory accuracy. The challenges of remote sensing image scene classification include (1) big intraclass diversity, (2) high interclass similarity (also known as low between-class separability),  (3) large variance of object/scene scales, and (4) coexistence of multiple ground objects, as shown in Fig. \ref{fig5}.
\\ \indent
In terms of within-class diversity, the challenge mainly stems from the large variations in the appearances of ground objects within the same semantic class. Ground objects commonly vary in style, shape, scale, and distribution, which makes it difficult to correctly classify the scene images.
For example, in Fig. \ref{fig5} (a), the churches appear in different building styles, and the airports and railway stations show in different shapes. In addition, when airborne or space platforms capture remote sensing images, there may be large differences in color and radiation intensity appearing within the same semantic class on account of the imaging conditions, which can be influenced by the factors such as weather, cloud, mist, etc. The variations in scene illumination may also cause within-class diversity, for example, the appearances of the scene labeled as ``beach'' show large differences under different imaging conditions, as shown in Fig. \ref{fig5} (a).
\begin{figure*}[!th]
	\centering
	\includegraphics[width=\linewidth]{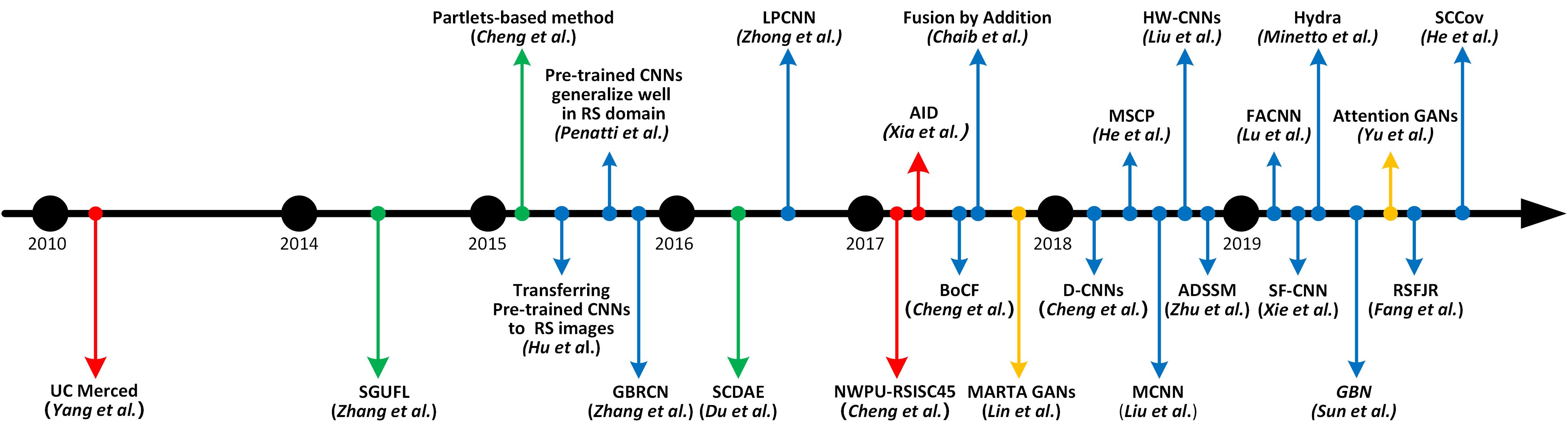}\\
	\caption{Milestones of deep learning-based remote sensing image scene classification, including different deep learning-based methods and data sets. The red line represents typical data sets. The green, blue, and orange lines stand for Autoencoder-based, CNN-based, and GAN-based remote sensing image scene classification, respectively.}
	\label{fig6}
\end{figure*}
\\ \indent
For between-class similarity, the challenge is chiefly caused by the presence of the same objects within different scene classes or the high semantic overlapping between scene categories. For instance, in Fig. \ref{fig5} (b), the scene classes of bridge and overpass both contain the same ground objects, namely bridge, and the basketball courts and tennis courts share high semantic information. Moreover, the ambiguous definition of scene classes degenerates inter-class dissimilarity. Some complex scenes are also similar with each other in terms of their visual contents. Therefore, it may be extremely difficult to distinguish these scene classes.
\\ \indent The large variance of object/scene scales is also a non-negligible challenge for remote sensing image scene classification. In remote sensing imaging, sensors operate at the orbits of various altitudes, from a few hundred kilometers to more than ten thousand kilometers, which leads to imaging altitude variation. With the examples illustrated in Fig. \ref{fig5} (c), the scenes of airplane, storage tank, and thermal power station have huge scale differences under different imaging altitudes. In addition, because of some intrinsic factors, the variations in size for each object/scene category can also exist, for example, the rivers shown in Fig. \ref{fig5} (c) are presented in several different sub-scenes---stream, brook, and creek. 
\\ \indent Moreover, owing to the complex and diverse distribution of ground objects and the wide bird’s-eye perspective of remote sensing imaging equipments, it is quite common that multiple ground objects appear in a single remote sensing image. As illustrated in Fig. \ref{fig5} (d), the scenes of commercial areas may contain buildings, cars, rivers, roads, parking lots, meadows, swimming pools, and playgrounds; roads, trees, bridges, rivers, and cars can coexist in the scenes of industrial areas; the scenes of ground track fields may accompany with the presence of swimming pools, cars, roads, meadows, and trees; the scenes of freeways contain meadows, trees, buildings, cars, rivers, bridges, forests, parking lots, etc. Faced with the situation, it is difficult for single-label remote sensing image scene classification to provide deep understanding for the contents of remote sensing images.

\section{Survey on Deep Learning-Based Remote Sensing Image Scene Classification Methods}
In the past decades, many researchers have committed to  scene classification of  remote sensing images, driven by its wide applications.  A number of advanced scene classification systems or approaches have been proposed,
especially driven by deep learning. Before deep learning came to the attention of this field, scene classification methods mainly relied on handcrafted features (e.g., Color Histogram (CH), texture descriptors (TD), GIST) or the representations generated by encoding local features via BoVW, IFK, SPM, etc. Later, considering that handcrafted features only extract low-level information, many researchers turned to looking at unsupervised learning methods (e.g., sparse coding, PCA, and k-means). By automatically learning discriminative features from unlabeled data, unsupervised learning-based methods have obtained good results in the scene classification of remote sensing images. Yet, unsupervised learning-based algorithms do not adequately exploit data class information, which limits their abilities to discriminate between different scene classes. Now, thanks to the availability of enormous labeled data, the advances in machine learning theory and the increased availability of computational resources, deep learning models (e.g., autoencoder, CNNs, and GANs) have shown powerful abilities to learn fruitful features and have permeated many research fields, including the area of remote sensing image scene classification. Currently, numerous deep learning-based scene classification algorithms have emerged and have yielded the best classification accuracy.
\begin{figure*}[!th]
	\centering
	\includegraphics[scale=0.75]{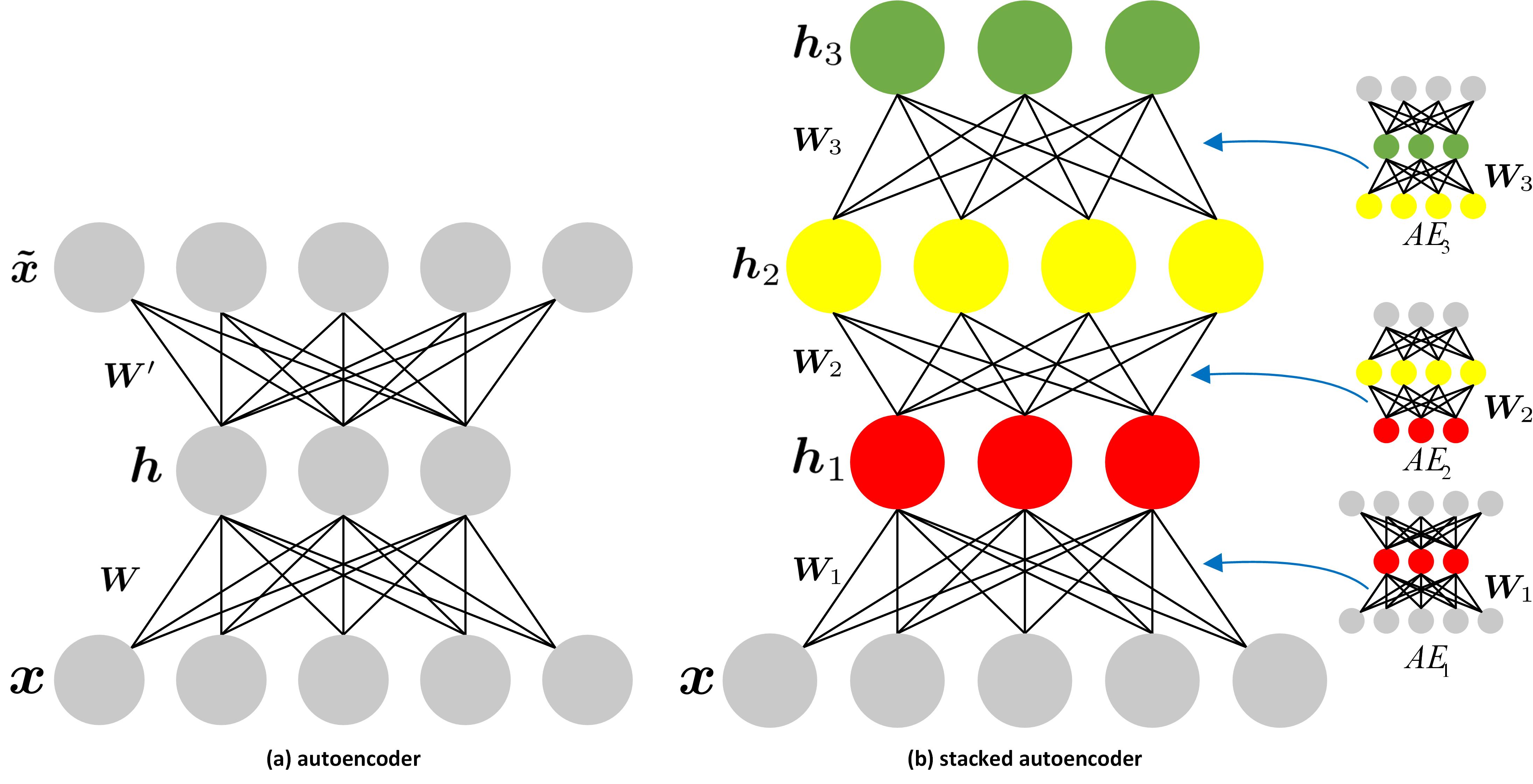}\\
	\caption{The architectures of (a) autoencoder and (b) stacked autoencoder. The red, yellow, and green nodes stand for the hidden layers of autoencoders $AE_{1}$, $AE_{2}$, and $AE_{3}$, respectively. When stacking these autoencoders, the output of the hidden layer of the previous autoencoder is the input of the following autoencoder. For example, the output of the hidden layer of $AE_{1}$ is the input of $AE_{2}$, and the output of the hidden layer of $AE_{2}$ is the input of $AE_{3}$.}
	\label{fig7}
\end{figure*}
In this section, we systematically survey about 50 deep learning-based algorithms for scene classification of remote sensing images. In Fig. \ref{fig6}, we present some milestone works. That is one small step for deep learning theory, but one giant leap for the scene classification of remote sensing images\cite{zhang2019remotely}. From autoencoder, to CNNs, and then to GANs, deep learning algorithms constantly update scene classification records. To sum up, most of the deep learning-based scene classification algorithms can be broadly divided into three main categories: autoencoder-based methods, CNN-based methods, and GAN-based methods. In what follows, we discuss the three categories of methods at great length.

\subsection{Autoencoder-Based Remote Sensing Image Scene Classification}
\subsubsection*{1) Brief introduction of autoencoder}
Autoencoder \cite{hinton2006reducing} is an unsupervised feature learning model, which consists of a sort of shallow and symmetrical neural network (see Fig. \ref{fig7} (a)). An autoencoder consists of three layers: input layer, hidden layer, and output layer. It contains two units---encoder and decoder. The transformation from input layer to hidden layer is the process of encoding. The process of encoding can be formulated as equation (\ref{1}), where $\bm{h} \in \mathbb{R}^{n}$ is the output of hidden layers, $f$ denotes a nonlinear mapping, $\boldsymbol{W} \in \mathbb{R}^{n \times m}$  stands for the encoding weight matrix, $\bm{x} \in \mathbb{R}^{m}$  denotes the input of autoencoder, and $\bm{b} \in \mathbb{R}^{n}$ is the bias vector. Decoding is the inverse of encoding, which is the transformation from hidden layer to output layer, and can be formulated as equation (\ref{2}), where $\bm{\tilde{x}} \in \mathbb{R}^{m}$  represents the reconstructed output, the decoding weight matrix is denoted by  $\boldsymbol{W}^{\prime} \in \mathbb{R}^{m\times n}$, and $\bm{b}^{\prime} \in \mathbb{R}^{m}$ stands for the bias vector.
\begin{equation}\label{1}
 \bm{h}=f(\boldsymbol{W} \cdot \bm{x}+\bm{b})
\end{equation}
\begin{equation}\label{2}
\tilde{\bm{x}}=f\left(\boldsymbol{W}^{\prime} \cdot \bm{h}+\bm{b}^{\prime}\right)
\end{equation}

Autoencoder is able to compress high dimensional features by minimizing the cost function that usually consists of a reconstruction error term and a regularization term. By using gradient descent with back propagation, autoencoder can learn the parameters of networks. In real applications, multilayer stacked autoencoders are used (see Fig. \ref{fig7} (b)) for feature learning. For example, three individual autoencoders $AE_{1}$, $AE_{2}$, and $AE_{3}$ are stacked together to form a stacked autoencoder, as shown in Fig. \ref{fig7} (b). When stacking these autoencoders, the output of the hidden layer of the previous autoencoder is the input of the following autoencoder. For example, the output of the hidden layer of $AE_{1}$ is the input of $AE_{2}$, and the output of the hidden layer of $AE_{2}$ is the input of $AE_{3}$. The key to training stacked autoencoders is how to initialize the network. The way of initializing the parameters of networks influences the network convergence especially the early layers, as well as the stability of training.
Fortunately, Hinton et al \cite{hinton2006reducing} provided a good solution to initialize the weight of the network by using restricted Boltzmann machines.
\begin{figure*}[t]
	\centering
	\includegraphics[scale=0.9]{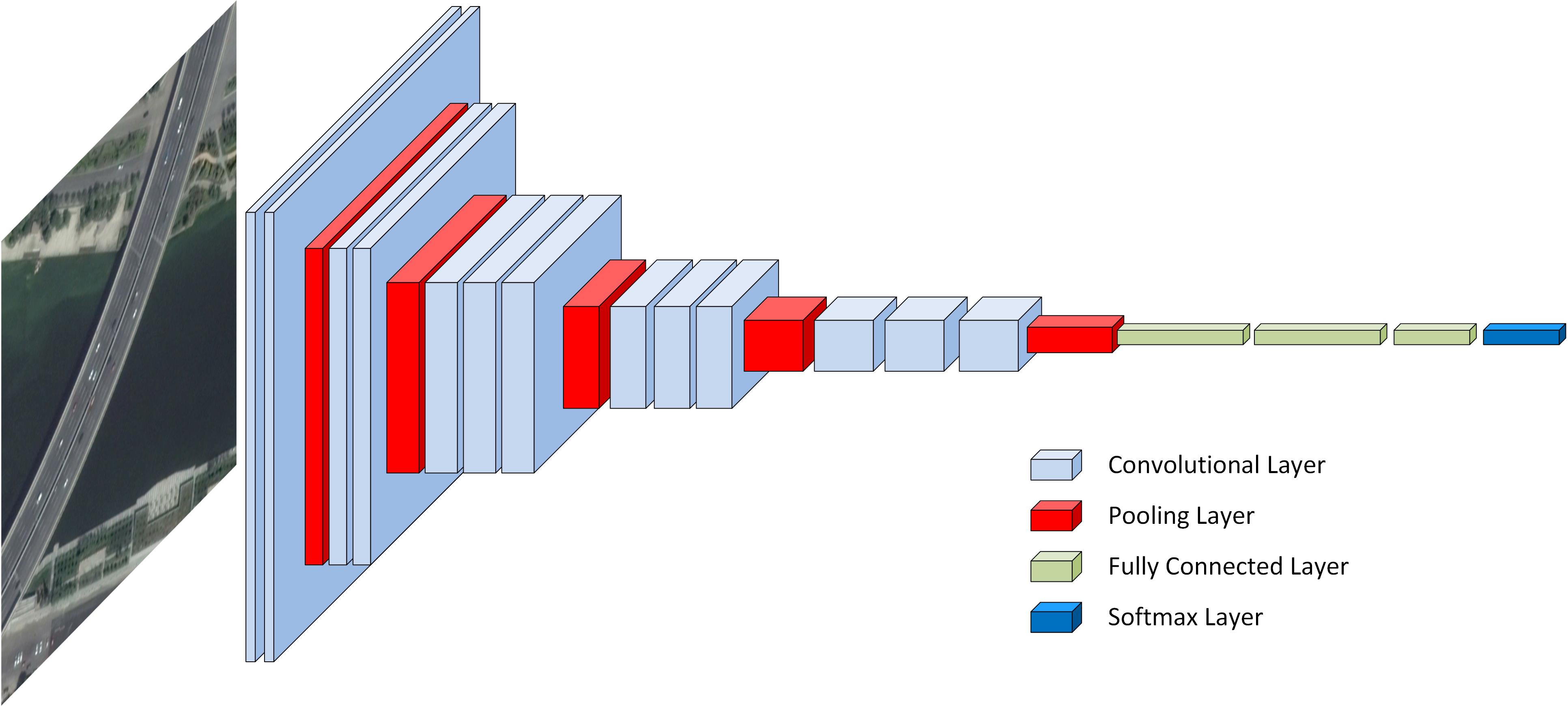}\\
	\caption{The architecture of CNNs.}
	\label{fig8}
\end{figure*}
\subsubsection*{2) Autoencoder-based scene classification methods}
Autoencoder is able to automatically learn mid-level visual representations from unlabeled data. The mid-level features plays an important role in remote sensing image scene classification before deep learning takes off in the remote sensing community. Zhang et al. \cite{zhang2014saliency} introduced sparse autoencoder to scene classification. Cheng et al. \cite{cheng2015learning} used the single-hidden-layer neural network and autoencoder for training more effective sparselets \cite{girshick2013discriminatively} to achieve efficient scene classification and object detection. In \cite{othman2016using}, Othman et. al proposed an remote sensing image scene classification algorithm relied on convolutional features and a sparse autoencoder. Han et al. \cite{han2017scene} provided the scene classification methods based on hierarchical convolutional sparse autoencoder. Cheng et al. \cite{cheng2015auto} demonstrated mid-level visual feature learned from autoencoder-based method is discriminative and able to facilitate scene classification tasks.
In light of the limitation of feature representation of a single autoencoder, some researchers stacked multiple autoencoders together. Du et al. \cite{du2016stacked} came up with stacked convolutional denoising autoencoder networks. After extensive experiments, their proposed framework showed superior classification performance. Yao et al. \cite{2016yao} integrated pairwise constraints into a stacked sparse autoencoder to learn more discriminative features for land-use scene classification and semantic annotation tasks.

{The autoencoder and the algorithms derived from autoencoder are unsupervised-learning methods and have obtained good results in scene classification of remote sensing images. However, most of the above-mentioned autoencoder-based methods cannot learn the best discrimination features to distinguish different scene classes because they do not fully exploit scene class information.}
\begin{figure*}[!th]
	\centering
	\includegraphics[scale=0.96]{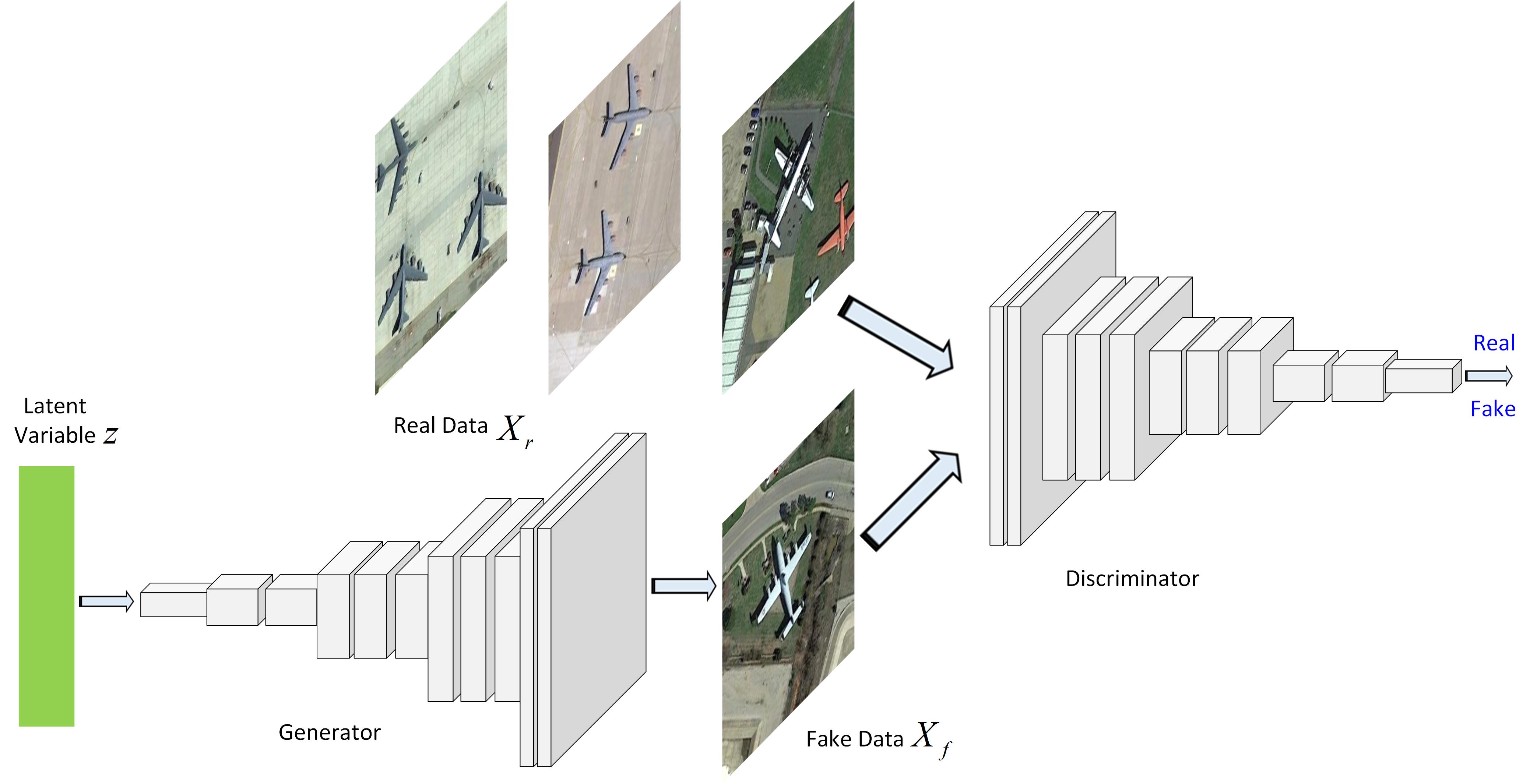}\\
	\caption{The architecture of GANs.}
	\label{fig9}
\end{figure*}

\subsection{CNN-Based Remote Sensing Image Scene Classification}
\subsubsection*{1) Brief introduction of CNN}
CNNs have shown powerful feature learning ability in the visual domain. Since Krizhevsky and Hinton proposed the Alexnet \cite{krizhevsky2012imagenet} in 2012, a deep CNN that obtained the best accuracy in the LSVRC, there have appeared an array of advanced CNN models, such as VGGNet \cite{simonyan2014very}, GoogleNet \cite{szegedy2015going}, ResNet \cite{he2016deep}, DensNet \cite{huang2017densely}, SENet \cite{hu2018squeeze}, and SKNet \cite{li2019selective}. CNNs are a kind of multi-layer network with learning ability that consists of convolutional layers, pooling layers, and fully connected layers (see Fig. \ref{fig8}).

(1)	Convolutional layers

Convolutional layers play an important role on feature extraction from images. The convolutional layer’s input  $\boldsymbol{X} \in \mathbb{R}^{n \times w \times h}$ consists of $n$  two-dimensional feature maps of size $w \times h$ . The output $\boldsymbol{H} \in \mathbb{R}^{m \times w^{\prime} \times h^{\prime}}$ of convolutional layers is $m$  two-dimensional feature maps of size $w^{\prime} \times h^{\prime}$  via convolutional kernels  $\boldsymbol{W}$. $\boldsymbol{W} \in \mathbb{R}^{m \times l \times l \times n}$  is $m$  trainable filters of size $l \times l \times n$  (typically $l=$1, 3 or 5). The entire process of convolution is described as equation (\ref{3}), where $*$  denotes two-dimensional convolution operation, additionally by using $\bm{b}$  to denote the $m$  dimensional bias term. In general, a non-linear activation function $f$  is performed after convolution operation. As the convolutional structure deepens, the convolutional layers can capture different level features (e.g., edges, lines, corners, structures, and shapes) from the input feature maps.
\begin{equation}\label{3}
\boldsymbol{H}=f(\boldsymbol{W} * \boldsymbol{X}+\bm{b})
\end{equation}

(2)	Pooling layers

Pooling layers are to execute a max or average operation over a small aera of each input feature map, which can be defined as equation (\ref{4}), where $pool$  represents the pooling function (e.g., average pooling, max pooling, and stochastic pooling), $\boldsymbol{H}_{l-1}$  and $\boldsymbol{H}_{l}$  denotes the input and output of the pooling layer respectively. Usually, pooling layers are applied between two successive convolutional layers.
Pooling operation can create invariance, such as small shifts and distortions. In the object detection and scene classification tasks, the characteristic of invariance provided by pooling layers is very important.
\begin{equation}\label{4}
\boldsymbol{H}_{l}=pool(\boldsymbol{H}_{l-1})
\end{equation}

(3)	Fully connected layers

Fully connected layers usually appear in the top layer of CNNs, which can summarize the features extracted from the bottom layers. Fully connected layers process its input $\tilde{\boldsymbol{X}}$ with linear transformation by weight $\tilde{\boldsymbol{W}}$ and bias $\tilde{\bm{b}}$, then map the output of linear transformation by a non-linear activation function $f$. The entire process can be formulated as equation (\ref{5}). In the task of classification, to output the probability of each class, a softmax classifier is connected to the last fully connected layer generally. The softmax classifier is used to normalize the fully connected layer output $\bm{y} \in \mathbb{R}^{c}$ ($c$ is the number of classes) between 0 and 1, which can be described as equation (\ref{6}), where $e$ is the exponential function. The output of softmax classifier denotes the probability that the given input image belongs to each class. The dropout method \cite{mekhalfi2015land} operates on the fully connected layers to avoid  overfitting because a fully connected layer usually contains a large number of parameters.
\begin{equation}\label{5}
\bm{y}=f(\tilde{\boldsymbol{W}}\cdot \tilde{\boldsymbol{X}}+\tilde{\bm{b}})
\end{equation}
\begin{equation}\label{6}
P\left(y_{i}\right)=\frac{e^{y_{i}}}{\sum_{i=1}^{c} e^{y_{i}}}
\end{equation}

\subsubsection*{2) CNN-based scene classification methods}
In the wake of CNNs successfully being applied to large-scale visual classification tasks, around 2015, the use of CNNs has finally taken off in the remote sensing image analysis field \cite{zhang2016deep,zhu2017deep}. Compared with traditional advanced methods, e.g., SIFT \cite{lowe2004distinctive}, HOG \cite{dalal2005histograms}, and BoVW \cite{yang2010bag}, CNNs have the advantage of end-to-end feature learning. Meanwhile, it can extract high-level visual features that handcrafted feature-based methods cannot learn. By using different strategies of exploiting CNNs, a variety of CNN-based scene classification methods \cite{zhang2015scene,zhong2016large,cheng2017bocf,yu2017unsupervised,liu2018scene,zhu2018deep,cheng2018deep} have emerged. Generally, the CNN-based methods of remote sensing image scene classification can be divided into three groups: using pre-trained CNNs as feature extractors, fine-tuning pre-trained CNNs on target data sets, and training CNNs from scratch.

(1) Using pre-trained CNNs as feature extractors

In the beginning, CNNs appeared as feature extractors. Penatti et al. \cite{penatti2015deep} introduced CNNs in 2015 into remote sensing image scene classification, and evaluated the generalization capability of off-the-shelf CNNs in classification of remote sensing images. Their experiments show that CNNs can obtain better results than low-level descriptors. Later, Hu et al. \cite{hu2015transferring} treated CNNs as feature extractors and investigated how to make full use of pre-trained CNNs for scene classification. In \cite{marmanis2015deep}, Marmanis et al. introduced a two-stage CNN scene classification framework. It used pre-trained CNNs to derive a set of representations from images. The extracted representations were then fed into shallow CNN classifiers. Chaib et al. \cite{chaib2017deep} fused the deep features extracted with VGGNet to enhance scene classification performance. In \cite{li2017integrating}, Li et al. fused pre-trained CNN features. The fused CNN features show better discrimination than raw CNN features in scene classification. Cheng et al. \cite{cheng2017bocf} designed the BoCF (bag of convolutional features) for remote sensing image scene classification by using off-the-shelf CNN features to replace traditional local descriptors such as SIFT. Yuan et al. \cite{yuan2018remote} rearranged the local features extracted by an already trained VGG19Net for remote sensing image scene classification. In \cite{he2018remote}, He et al. proposed a novel multilayer stacked covariance pooling algorithm (MSCP) for remote sensing image scene classification. MSCP can combine multilayer feature maps extracted from pre-trained CNN automatically. Lu et al. \cite{lu2019feature} introduced an feature aggregation CNN (FACNN) for scene classification. FACNN learns scene representations through exploring semantic label information. These methods all used pre-trained CNNs as feature extractors and then fused or combined the features extracted by existing CNNs. It is worth noticing that the strategy of using off-the-shelf CNNs as feature extractors is simple and effective on small-scale data sets.

(2) Fine-tuning pre-trained CNNs

However, when the amount of training samples is not adequate to train a new CNN from scratch, fine-tuning an already trained CNNs on target data sets is a good choice. Castelluccio et al. \cite{castelluccio2015land} delved into the use of CNNs for remote sensing image scene classification by experimenting with three learning approaches: using pre-trained CNNs as feature extractors, fine tuning, and training from scratch. And they concluded that fine-tuning gave better results than full training when the scale of data sets is small. This made researchers interested in fine-adjusting scene classification networks or optimizing its loss functions. Cheng et al. \cite{cheng2018deep} designed a novel objective function for learning discriminative CNNs (D-CNNs). The D-CNNs shows better discriminability in scene classification. In \cite{liu2018hw}, Liu et al. coupled CNN with a hierarchical Wasseratein loss function (HW-CNNs) to improve CNN’s discriminatory ability. Minetto et al. \cite{minetto2019hydra} devised a new remote sensing image scene classification framework, named Hydra, which is an ensembles of CNNs and achieves the best results on the NWPU-RESISC45 data set. Wang et al. \cite{wang2018scene} introduced attention mechanism into CNNs and designed the ARCNet  ( attention recurrent convolutional network ) for scene classification. It is capable of highlighting key areas and discard noncritical information. In \cite{liu2018m}, to handle the problem of object scale variation in scene classification, Liu et al. formulated the multiscale CNN (MCNN). Fang et al. \cite{fang2019robust} designed a robust space-frequency joint representation (RSFJR) for scene classification by adding a frequency domain branch to CNNs. Because of fusing features from the space and frequency domains, the proposed method is able to provide more discriminative feature representations. Xie et al. \cite{xie2019scale} designed a scale-free CNN (SF-CNN) for the task of scene classification. SF-CNN can accept the images of arbitrary size as input without any resizing operation. Sun et al. \cite{sun2019remote} proposed a gated bidirectional network (GBN) for scene classification, which can get rid of the interference information and aggregate the interdependent information among different CNN layers.In the above-mentioned methods, CNNs can learn discriminative features and obtain better performance  by fine adjusting their structures, optimizing their objective function, or fine-tuning the modified CNNs on the target data sets.

(3) Training CNNs from scratch

Even though fine-tuning pre-trained CNNs can achieve remarkable performance, there exist some limitations relying on pre-trained CNNs: learned features are not fully suitable for the characteristics of target data sets and it is inconvenient for researchers to modify pre-trained CNNs. In \cite{chen2018training}, Chen et al. introduced knowledge distillation into scene classification to boost the performance of light CNNs. Zhang et al. \cite{zhang2019lightweight} illustrated a lightweight and effective CNN that introduces the dilated convolution and channel attention into Mobilenetv2 \cite{sandler2018mobilenetv2} for scene classification. In addition, it is of considerable interest to design more effective and robust CNNs for scene classification. He et al.\cite{he2019skip} introduced a novel skip-connected covariance (SCCov) network for remote sensing image scene classification. The SCCov is to add skip connection and covariance pooling to CNNs, which can reduce the amount of parameters and achieve better classification performance. In \cite{zhang2015scene}, Zhang et al. presented a gradient boosting random convolutional network (GBRCN) for scene classification via assembling different deep neural networks.
\\ \indent These CNN-based methods have obtained astonishing scene classification results. However, they generally require numerous annotated samples to fine-tune already trained CNNs or train a network from scratch.

\subsection{GAN-Based Remote Sensing Image Scene Classification}
\subsubsection*{1) Brief introduction of GAN}
Generative Adversarial Network (GAN) \cite{goodfellow2014generative} is another important and promising machine learning method. As its name implies, GAN models the distribution of data via adversarial learning based on a minimax two-player game, and generates real-like data. GANs contain a pair of components---the discriminator $\mathrm{D}$ and generator $\mathrm{G}$. As shown in Fig. \ref{fig9}, $\mathrm{G}$ can be analogues to a group of counterfeiters who take the role of generating fake currency, while $\mathrm{D}$ can be thought of as polices who determine whether the currency is made by $\mathrm{G}$ or bank. $\mathrm{G}$ and $\mathrm{D}$ constantly pit against each other in this game until $\mathrm{D}$ cannot distinguish between the counterfeit currency and genuine articles. GANs see the competition between $\mathrm{G}$ and $\mathrm{D}$ as the sole training criterion. $\mathrm{G}$ takes an input $\bm{z}$, which is a latent variable obeying a prior distribution $p_{\bm{z}}(\bm{z})$, then maps $\bm{z}$ with noise into data space by using a differential function $G\left(\bm{z} ;\bm{\theta}_{g}\right)$, where $\bm{\theta}_{g}$ denotes the generator $\mathrm{G}$'s parameters.
$\mathrm{D}$ outputs the probability of the input data $\bm{x}$ that comes from real data rather than generator through a mapping $D\left(\bm{x} ;\bm{\theta}_{d}\right)$ with parameters $\bm{\theta}_{d}$, where $\bm{\theta}_{d}$  denotes the discriminator $\mathrm{D}$'s parameters.
The entire process of the two-player minimax game is described as equation (\ref{6}), where $p_{data}$ is the distribution of data $\bm{x}$ and $\mathrm{V}(G, D)$ is an object function.
From $\mathrm{D}$'s perspective, given an input data generated by $\mathrm{G}$, $\mathrm{D}$ will play a role in minimizing its output. While if a sample is real data, $\mathrm{D}$ will maximize its output. This is the reason why the term $\log (1-D(G(\bm{z})))$  is plugged into equation (\ref{6}). Meanwhile, to fool $\mathrm{D}$, $\mathrm{G}$ makes an effort to maximize  $\mathrm{D}$'s output when a generated data is input to $\mathrm{D}$. Thus the relationship that $\mathrm{D}$ wants to $\text { maximize } \mathrm{V}(G, D)$ and $\mathrm{G}$ struggles to $\text { minimize } \mathrm{V}(G, D)$ is formed.
\begin{equation}\label{6}
\begin{split}
\min _{G} \max _{D} V(G, D)=E_{\bm{x} \sim p_{\text {data}}(\bm{x})}[\log D(\bm{x})] \\
+E_{\bm{z} \sim p_{t}(\bm{z})}[\log (1-D(G(\bm{z})))]
\end{split}
\end{equation}

\begin{table*}[!th]
	\centering
    \vspace{\baselineskip}
	\caption{13 publicly available data sets for remote sensing image scene classification.}\label{table2}
	\resizebox{0.98\textwidth}{!}{
		\renewcommand\arraystretch{1.8}
		\begin{tabular}{|p{3.2cm}<{\centering}|p{2cm}<{\centering}|p{1.9cm}<{\centering}|p{1.8cm}<{\centering}|p{2.2cm}<{\centering}|p{0.8cm}<{\centering}|p{0.8cm}<{\centering}|p{2.5cm}<{\centering}|p{1cm}<{\centering}|}
			\hline
			Data sets &  \multicolumn{1}{m{2cm}<{\centering}|}{Image number per class} & \multicolumn{1}{m{1.9cm}<{\centering}|}{Number of scene classes }& \multicolumn{1}{m{1.8cm}<{\centering}|}{Total image number} & Image size & \multicolumn{2}{c|}{Training ratios}& Data sources &Year\\
			\hline
			UC Merced \cite{yang2010bag} & 100
			& 21 & 2100
			&256 $\times$ 256 &50$\%$ &80$\%$ &Aerial orthoimagery &2010\\
			\hline

			WHU-RS19 \cite{xia2010structural} &50$\sim$61
			& 19 & 1005
			& 600$\times$600 &40$\%$ &60$\%$ &Google Earth &2012\\
			
            \hline
			RSSCN7 \cite{zou2015deep} & 400
			& 7 & 2800
			& 400$\times$400 &20$\%$ &50$\%$ &Google Earth&2015\\
			\hline

			Brazilian Coffee Scene \cite{penatti2015deep} & 1438
			& 2 & 2876
			& 64$\times$64 & \multicolumn{2}{c|}{50$\%$} &SPOT sensor &2015\\
            \hline
            SAT-4/-6 \cite{basu2015} & 125000/67500
			& 4/6 & 500000/405000
			& 28$\times$28  & \multicolumn{2}{c|}{80$\%$} &\multicolumn{1}{m{2.5cm}<{\centering}|}{National Agriculture Imagery Program}&2015\\
            \hline

			SIRI-WHU \cite{zhao2015dirichlet} & 200
			& 12 & 2400
			& 200$\times$200 & \multicolumn{2}{c|}{50$\%$} &Google Earth &2016\\
			
			\hline
			RSC11 \cite{zhao2016feature} & about 100
			& 11 & 1232
			& 512$\times$512 & \multicolumn{2}{c|}{50$\%$} &Google Earth &2016\\
			
			\hline		
			AID \cite{xia2017aid} & 220$\sim$420
			& 30 & 10000
			& 600$\times$600 &20$\%$ &50$\%$ &Google Earth  &2017\\
			
			\hline
			NWPU-RESISC45 \cite{cheng2017remote} & 700
			& 45 & 31500
			& 256$\times$256 &10$\%$ &20$\%$ &Google Earth  &2017\\
			
            \hline
			RSI-CB128/-CB256 \cite{li2017rsi} & about 800/690
			& 45/35 & 36000/24000
			&128$\times$128/256$\times$256 &50$\%$ &80$\%$ &\multicolumn{1}{m{2.5cm}<{\centering}|}{Google Earth \& Bing Maps} &2017\\
			\hline

			OPTIMAL-31 \cite{wang2018scene} & 60
			& 31 & 1860
			& \multicolumn{1}{m{2cm}<{\centering}|}{256$\times$256} &\multicolumn{2}{c|}{80$\%$} &Google Earth &2018\\
			\hline

			EuroSAT \cite{helber2019eurosat} & 2000$\sim$3000
			& 10 & 27000
			& 64$\times$64 &\multicolumn{2}{c|}{80$\%$} &Sentinel-2  &2019\\
			\hline

			BigEarthNet \cite{sumbul2019bigearthnet} & 328$\sim$217119
			& 44 & 590326
			& 120$\times$120 &\multicolumn{2}{c|}{60$\%$} &Sentinel-2 &2019\\
			\hline
	\end{tabular}}
\end{table*}

\begin{figure*}[!th]
	\centering
    \vspace{\baselineskip}
    \vspace{\baselineskip}
	\includegraphics[width=\linewidth]{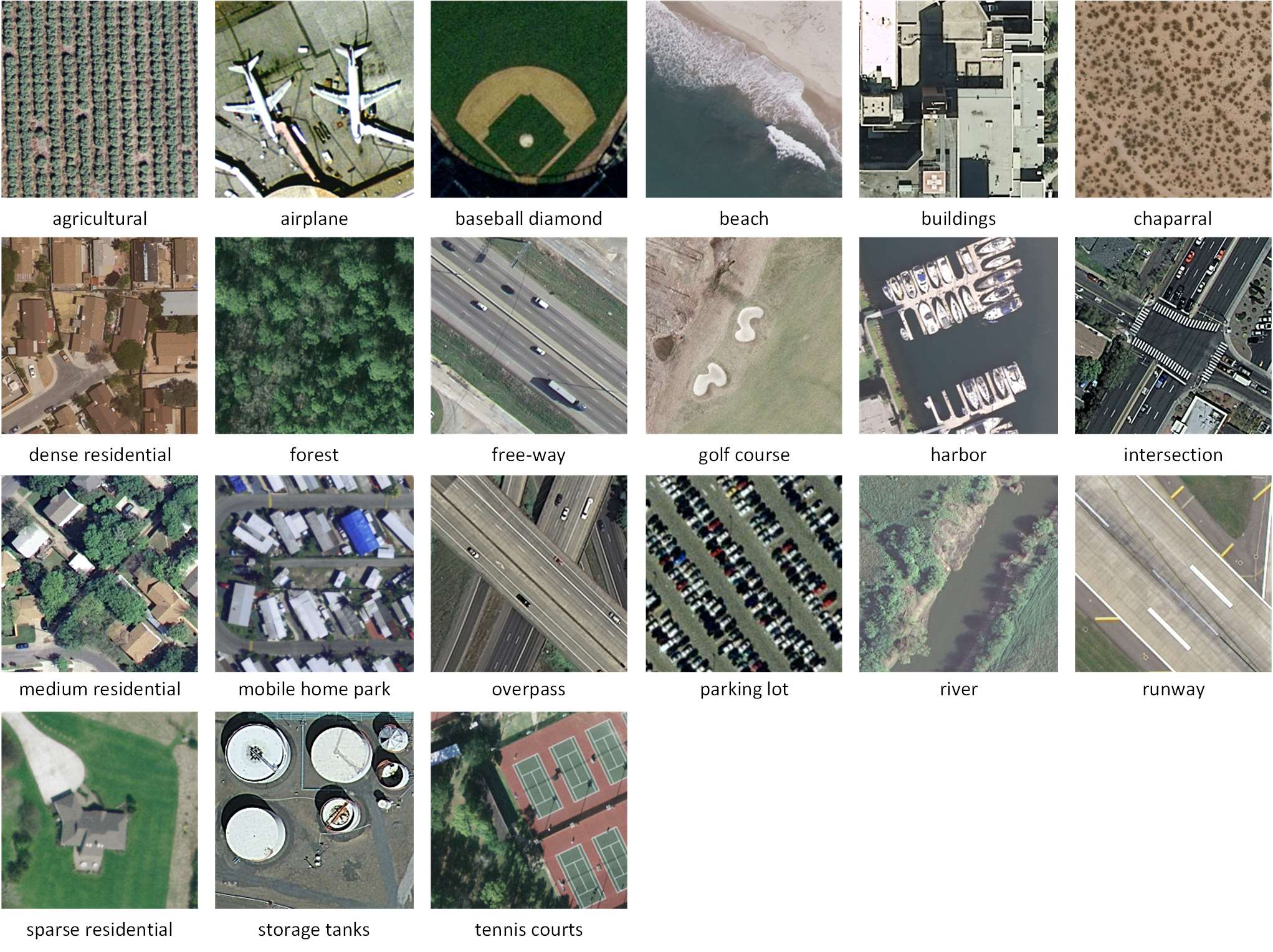}\\
	\caption{Some example images from the UC-Merced  data set.}
	\label{fig10}
\end{figure*}
\begin{figure*}[th]
	\centering
	\includegraphics[width=\linewidth]{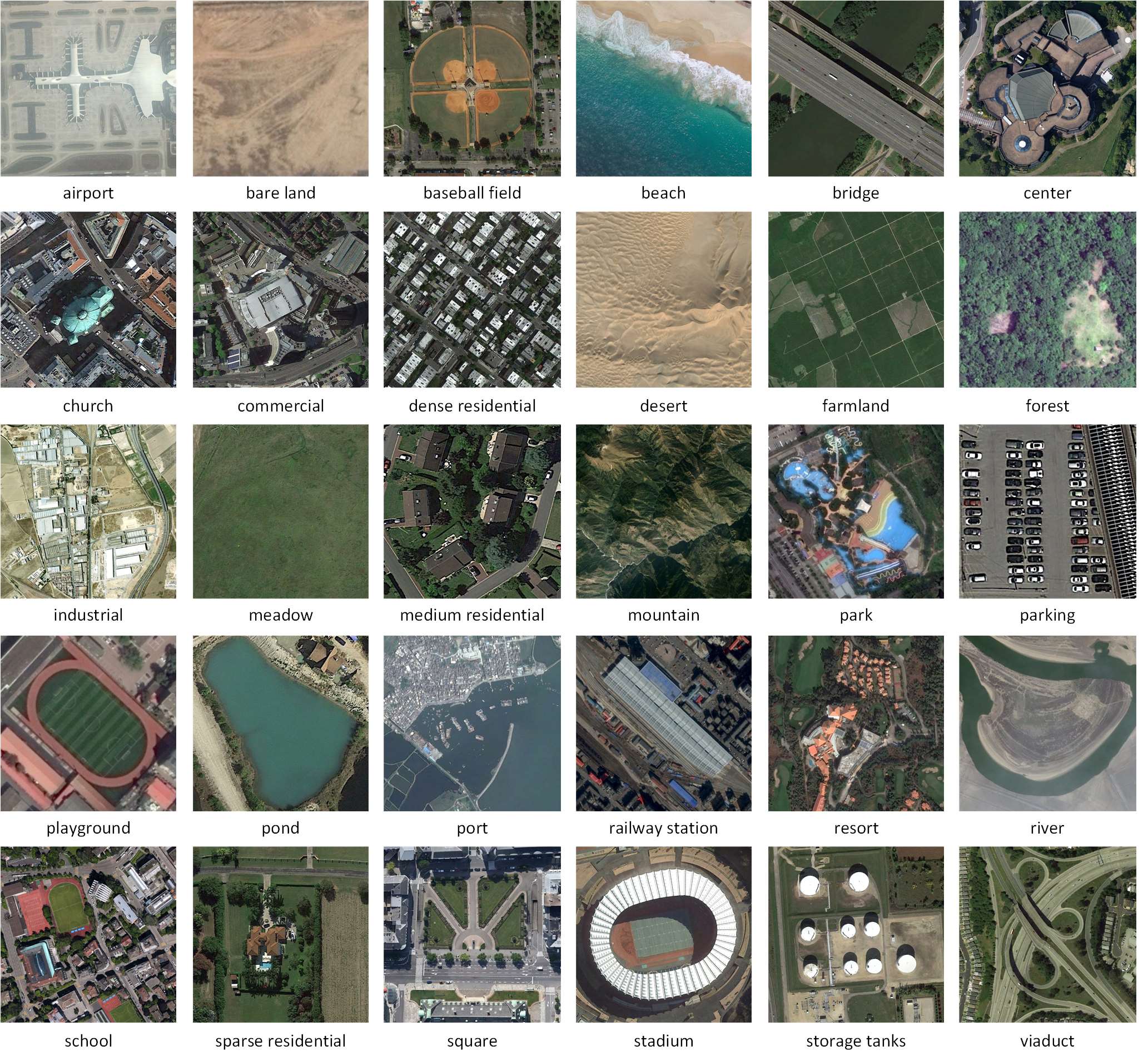}\\
	\caption{Some example images from the AID data set.}
	\label{fig11}
\end{figure*}
\begin{figure*}[!th]
	\centering
	\includegraphics[width=\linewidth]{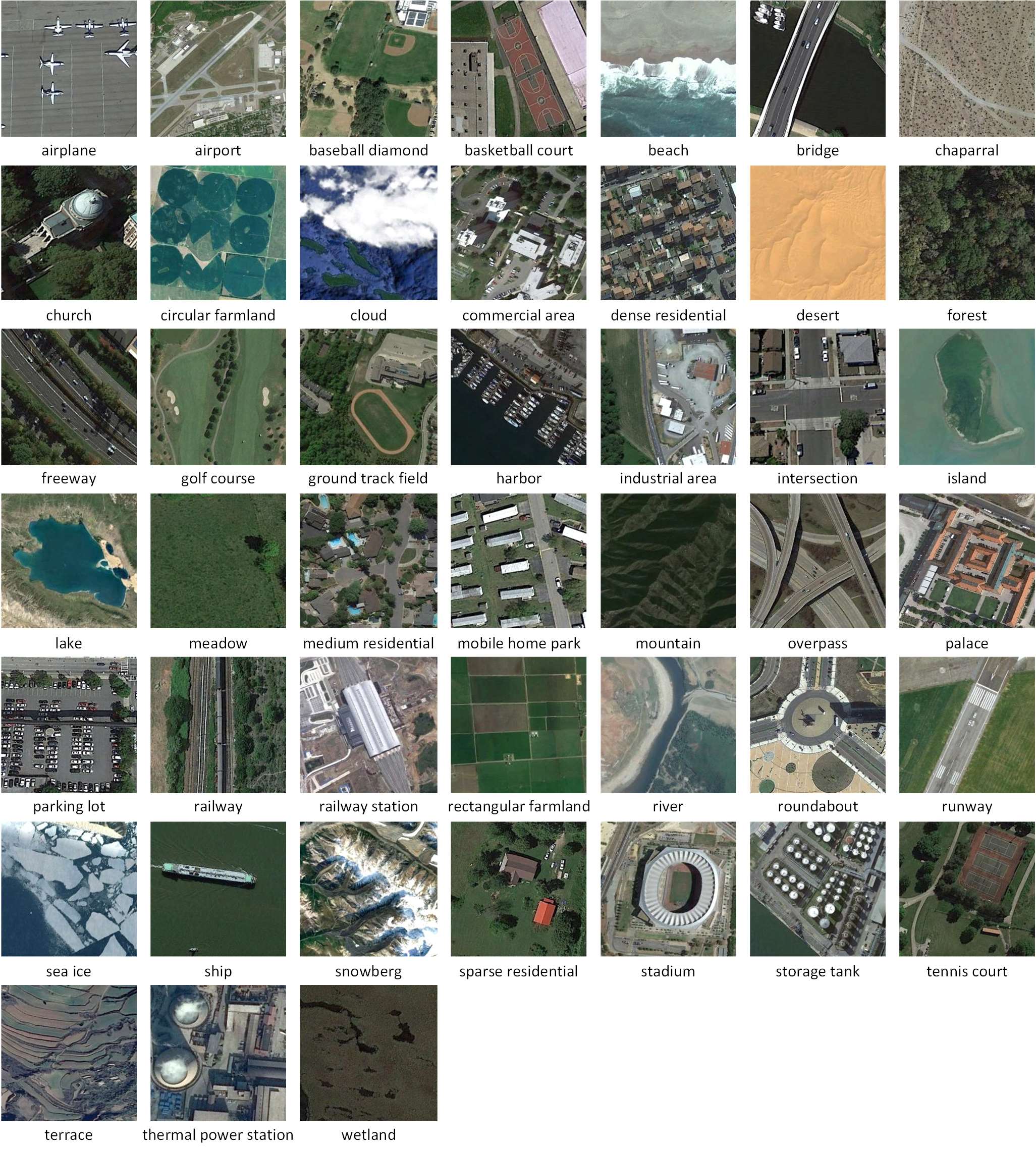}\\
	\caption{Some example images from the NWPU-RESISC45 data set.}
	\label{fig12}
\end{figure*}

\subsubsection*{2) GAN-based scene classification methods}
As a key method for unsupervised learning, since the introduction by Goodfellow et al. \cite{goodfellow2014generative} in 2014, GANs have been gradually applied to many tasks such as image to image translation, sample generation, image super-resolution, and so on. Facing the tremendous volume of remote sensing images, CNN-based methods need to use massive labeled samples to train models. However, annotating samples is labor-intensive. Some researchers began to employ GANs to scene classification. In 2017, Lin et al. \cite{lin2017marta} proposed a multiple-layer feature-matching generative adversarial networks (MARTA GANs) for the task of scene classification. Duan et al. \cite{duan2018gan} used an adversarial net to assist in mining the inherent and discriminative features from remote sensing images. The dug features are able to enhance the classification accuracy. Bashmal et al. \cite{bashmal2018siamese} provided a GAN-based method, called Siamese-GAN, to handle the aerial vehicle images classification problems under  cross-domain conditions. In \cite{xu2018remote}, to generate high-quality remote sensing images for scene classification, Xu et al. added the scaled exponential linear unites to GANs. Ma et al. \cite{ma2019siftinggan} designed the SiftingGAN, which can generate a large variety of authentic annotated samples for scene classification. Teng et al. \cite{teng2019classifier} presented a classifier-constrained adversarial network for cross-domain semi-supervised scene classification. Han et al. \cite{han2018semi} introduced a generative framework, named SSGF, to scene classification. Yu et al. \cite{yu2019attention} devised an attention GAN for scene classification. Attention GAN achieves better scene classification performance by enhancing the representation power of the discriminator.

In the area of remote sensing scene image classification, most of GAN-based methods usually use GANs for sample generation or feature learning in an adversarial manner. Compared with CNN-based scene classification methods, only a small number of literatures about GAN-based scene classification method have been reported so far, and the performance of GAN-based scene classification is inferior to CNN-based methods. In addition, most of GAN-based scene classification methods cannot be trained end-to-end because they often require labels for training an additional classifier. However, the powerful self-supervised feature learning capacity of GANs provides a promising future direction for scene classification.

\section{Survey on Remote Sensing Image Scene Classification Benchmarks}
Data sets play an irreplaceable role on the advance of scene classification. Meanwhile, they are crucial for developing and evaluating various scene classification methods. As the number of high-resolution remote sensing sensors increases, the access to massive high-resolution remote sensing images makes it possible to build large-scale scene classification benchmarks. In the past few years, the researchers from different groups have proposed several publicly available high-resolution benchmark data sets for scene  classification of remote sensing images \cite{yang2010bag,xia2010structural,zhao2015dirichlet,basu2015,zou2015deep,zhao2016feature,li2017rsi,cheng2017remote,xia2017aid,wang2018scene,sumbul2019bigearthnet,penatti2015deep,helber2019eurosat} to facilitate this field forward.
Starting with the UC-Merced data set \cite{yang2010bag}, some representative data sets include WHU-RS19 \cite{xia2010structural}, SAT-4$\&6$ \cite{basu2015}, RSSCN7 \cite{zou2015deep}, Brazilian Coffee Scene \cite{penatti2015deep}, RSC11 \cite{zhao2016feature}, SIRI-WHU \cite{zhao2015dirichlet}, RSCI-CB \cite{li2017rsi}, AID \cite{xia2017aid}, NWPU-RESISC45 \cite{cheng2017remote}, OPTIMAL-31 \cite{wang2018scene}, EuroSAT \cite{helber2019eurosat}, and BigEarthNet \cite{sumbul2019bigearthnet}. The characteristics of these 13 data sets are listed in Table \ref{table2}. Among them, the UC-Merced data \cite{yang2010bag}, AID data set \cite{xia2017aid}, and NWPU-RESISC45 data set \cite{cheng2017remote} are three commonly-used benchmark data sets, which will be introduced below in detail.

\subsection{UC-Merced Data Set}
The UC-Merced data set\footnote{\url{http://weegee.vision.ucmerced.edu/datasets/form.html}}\cite{yang2010bag} was released in 2010 and contains 21  scene classes. Each category consists of 100 land-use images. In total, the data set comprises 2100 scene images, of which the pixel resolution is 0.3 m. These images were obtained from United States Geological Survey National Map of 21 U.S. regions and fixed at $256\times256$ pixels. Fig. \ref{fig10} lists the samples of each category from the data set. Up to now, the data set continues to be broadly employed for scene classification. When conducting algorithm evaluation, two widely-used training ratios are 50$\%$ and 80$\%$, and the remaining 50$\%$ and 20$\%$ are used for testing.

\subsection{AID Data Set}
The AID \cite{xia2017aid} data set\footnote{\url{www.lmars.whu.edu.cn/xia/AID-project.html}} is a relatively large-scale data set for aerial scene classification. It was published in 2017 by Wuhan University and consists of 30 scene classes. Each scene class consists of 220 to 420 images, which were cropped from Google Earth imagery and fixed at $600\times600$ pixels. In total, the data set comprises 10000 scene images. Fig. \ref{fig11} lists the samples of each category from the data set. Different from the UC-Merced data set, the AID data set is multi-sourced because these aerial images were captured with different sensors. Moreover, the data set is also multi-resolution and the pixel resolution of each scene categories varies from about 8 m to about 0.5 m. When conducting algorithm evaluation, two widely-used training ratios are 20$\%$ and 50$\%$, and the remaining 80$\%$ and 50$\%$ are used for testing.

\subsection{NWPU-RESISC45 Data Set}
To the best of our knowledge, the NWPU-RESISC45 data set\footnote{\url{http://www.escience.cn/people/gongcheng/NWPU-RESISC45.html}}\cite{cheng2017remote}, released by Northwest Polytechnical University, is currently the largest scene classification data set. It consists of 45 scene categories. Each category consists of 700 images, which were obtained from Google Earth and fixed at $256\times256$ pixels. In total, the data set comprises 31500 scene images, which is chosen from more than 100 countries and regions. Apart from some specific classes with lower spatial resolution (e.g., island, lake, mountain, and iceberg), the pixel resolution of most the scene categories varies from about 30 m to 0.2 m. Fig. \ref{fig12}  lists the samples of each category from the data set. The release of NWPU-RESISC45 data set has allowed deep learning models to develop their full potential. When conducting algorithm evaluation, two widely-used training ratios are 10$\%$ and 20$\%$, and the remaining 90$\%$ and 80$\%$ are used for testing.

\begin{table*}[!th]
	\centering
	\caption{Overall accuracy ($\%$) comparison of 21  scene classification methods on the UC-Merced data set. }\label{table3}
	\resizebox{0.95\textwidth}{!}{
		\renewcommand\arraystretch{1.6}
		\begin{tabular}{|p{3.8cm}<{\centering}|p{3.8cm}<{\centering}|p{1.2cm}<{\centering}|p{2.5cm}<{\centering}|p{2.0cm}<{\centering}|p{2.0cm}<{\centering}|}	
			\hline
			\multicolumn{2}{|c|}{ \multirow{2}*{Method} }&\multirow{2}*{Year} & \multirow{2}*{Publication} &\multicolumn{2}{c|}{Training ratio} \\
			\cline{5-6}
			\multicolumn{2}{|c|}{} & & &50$\%$ &80$\%$\\
			\hline
			\multirow{3}*{Autoencoder-based}  &SGUFL \cite{zhang2014saliency} &2014 &IEEE TGRS &- &82.72$\pm$1.18\\
			\cline{2-6}
			&partlets-based method \cite{cheng2015effective} &2015 &IEEE TGRS &88.76$\pm$0.79 &-\\
			\cline{2-6}
			&SCDAE \cite{du2016stacked} &2016 &IEEE TCYB &- &93.7$\pm$1.3\\
			\hline
			
			\multirow{15}*{CNN-based}  &GBRCN \cite{zhang2015scene}
			&2015	&IEEE TGRS	&-	&94.53 \\
			\cline{2-6}
			&LPCNN \cite{zhong2016large}
			&2016	&JARS	&-	&89.90 \\
			\cline{2-6}
			&Fusion by Addition \cite{chaib2017deep}
			&2017	&IEEE TGRS	&-	&97.42$\pm$1.79 \\
			\cline{2-6}
			&ARCNet-VGG16 \cite{wang2018scene}
			&2018	&IEEE TGRS	&96.81$\pm$0.14	&99.12$\pm$0.40 \\
			\cline{2-6}
			&MSCP \cite{he2018remote}
			&2018	&IEEE TGRS	&-	&98.36$\pm$0.58 \\
			\cline{2-6}
			&D-CNNs \cite{cheng2018deep}
			&2018	&IEEE TGRS	&-	&98.93$\pm$0.10 \\
			
			\cline{2-6}
			&MCNN \cite{liu2018m}
			&2018	&IEEE TGRS	&-	&96.66$\pm$0.9 \\
			\cline{2-6}
			&ADSSM \cite{zhu2018adaptive}
			&2018	&IEEE TGRS	&-	&99.76$\pm$0.24  \\
			
			\cline{2-6}
			&FACNN \cite{lu2019feature}
			&2019	&IEEE TGRS	&-	&98.81$\pm$0.24 \\
			
			\cline{2-6}
			&SF-CNN \cite{xie2019scale}
			&2019	&IEEE TGRS	&-	&99.05$\pm$0.27 \\
			
            \cline{2-6}
			&SCCov \cite{he2019skip}
			&2019	&IEEE TNNLS	&-	&99.05$\pm$0.25 \\
			
			\cline{2-6}
			&RSFJR \cite{fang2019robust}
			&2019	&IEEE TGRS	&97.21$\pm$0.65	 &- \\
			
			\cline{2-6}
			&GBN \cite{sun2019remote}
			&2019	&IEEE TGRS	&97.05$\pm$0.19	&98.57$\pm$0.48 \\
			
			\cline{2-6}
			&ADFF \cite{zhu2019attention}
			&2019	&Remote Sensing	&96.05$\pm$0.56	&97.53$\pm$0.63 \\
			
			\cline{2-6}
			&CNN-CapsNet \cite{zhang2019remote}
			&2019	&Remote Sensing	&97.59$\pm$0.16	&99.05$\pm$0.24 \\
			
			\cline{2-6}
			&Siamese ResNet50 \cite{liu2019siamese}
			&2019	&IEEE GRSL	&90.95	&94.29    \\
			\hline
			
			\multirow{2}*{GAN-based}
			&MARTA GANs \cite{lin2017marta}
			&2017	&IEEE GRSL	&85.5$\pm$0.69	&94.86$\pm$0.80 \\
			
			\cline{2-6}
			&Attention GANs \cite{yu2019attention}
			&2019	&IEEE TGRS	&89.06$\pm$0.50	&97.69$\pm$0.69 \\
			\hline
	\end{tabular}}
\end{table*}

\begin{table*}[!th]
	\centering
	\caption{Overall accuracy ($\%$) comparison of 16 scene classification methods on the AID data set.}\label{table4}
	\resizebox{0.95\textwidth}{!}{
		\renewcommand\arraystretch{1.6}
		\begin{tabular}{|p{3.8cm}<{\centering}|p{3.8cm}<{\centering}|p{1.2cm}<{\centering}|p{2.5cm}<{\centering}|p{2.0cm}<{\centering}|p{2.0cm}<{\centering}|}	
			\hline
			\multicolumn{2}{|c|}{ \multirow{2}*{Method} }&\multirow{2}*{Year} & \multirow{2}*{Publication} &\multicolumn{2}{c|}{Training ratio} \\
			\cline{5-6}
			\multicolumn{2}{|c|}{} & & &20$\%$ &50$\%$\\
			\hline
			
			\multirow{14}*{CNN-based}
			&Fusion by Addition \cite{chaib2017deep}
			&2017	&IEEE TGRS	&-	&91.87$\pm$0.36  \\
			
			\cline{2-6}
			&ARCNet-VGG16 \cite{wang2018scene}
			&2018	&IEEE TGRS	&88.75$\pm$0.40	&93.10$\pm$0.55 \\
			
			\cline{2-6}
			&MSCP \cite{he2018remote}
			&2018	&IEEE TGRS	&91.52$\pm$0.21	&94.42$\pm$0.17  \\
			
			\cline{2-6}
			&D-CNNs \cite{cheng2018deep}
			&2018	&IEEE TGRS	 &90.82$\pm$0.16	&96.89$\pm$0.10\\
			
			\cline{2-6}
			&MCNN \cite{liu2018m}
			&2018	&IEEE TGRS	&-	&91.80$\pm$0.22 \\
			
			\cline{2-6}
			&HW-CNNs \cite{liu2018hw}
			&2018	&IEEE TGRS	&-	&96.98$\pm$0.33 \\
			
			\cline{2-6}
			&FACNN \cite{lu2019feature}
			&2019	&IEEE TGRS	&-	&95.45$\pm$0.11 \\
			
			\cline{2-6}
			&SF-CNN \cite{xie2019scale}
			&2019	&IEEE TGRS &93.60$\pm$0.12	&96.66$\pm$0.11\\
			
            \cline{2-6}
			&SCCov \cite{he2019skip}
			&2019	&IEEE TNNLS &93.12$\pm$0.25	&96.10$\pm$0.16\\

             \cline{2-6}
			&CNNs-WD \cite{WD2019}
			&2019	&IEEE GRSL &-	&97.24$\pm$0.32\\

			\cline{2-6}
			&RSFJR \cite{fang2019robust}
			&2019	&IEEE TGRS	&-	 &96.81$\pm$1.36 \\
			
			\cline{2-6}
			&GBN \cite{sun2019remote}
			&2019	&IEEE TGRS	 &92.20$\pm$0.23 &95.48$\pm$0.12 \\
			
			\cline{2-6}
			&ADFF \cite{zhu2019attention}
			&2019	&Remote Sensing	 &93.68$\pm$0.29	&94.75$\pm$0.25\\
			
			\cline{2-6}
			&CNN-CapsNet \cite{zhang2019remote}
			&2019	&Remote Sensing	  &93.79$\pm$0.13	&96.32$\pm$0.12\\
			
			\hline
			\multirow{2}*{GAN-based}
			&MARTA GANs \cite{lin2017marta}
			&2017	&IEEE GRSL	 &75.39$\pm$0.49	&81.57$\pm$0.33 \\
			
			\cline{2-6}
			&Attention GANs \cite{yu2019attention}
			&2019	&IEEE TGRS	&78.95$\pm$±0.23	&84.52$\pm$0.18  \\
			\hline
	\end{tabular}}
\end{table*}
\section{Performance Comparison and Discussion}
\subsection{Evaluation Criteria}
There exist three commonly-used criteria for evaluating the performance of the task of remote sensing image scene classification: overall accuracy (OA), average accuracy (AA), and confusion matrix. The metric of OA is an evaluation of the performance of the classifiers over the entire test data set, which is formulated as the total number of accurately classified samples $N_{c}$ divided by the total number of tested samples $N_{t}$, as described in equation (\ref{7}). OA is a commonly-used criterion for evaluating the performance of the methods for scene classification of remote sensing images. The criterion of AA is defined as the sum of the accuracies of each category $A_{i}$ divided by the total number of class $c$, as described in equation (\ref{8}). When the sample number of each category is equal on the test set, OA and AA have the same value. The confusion matrix is a detailed classification result table about the performance of each single classifier. For each element $x_{ij}$ in the table, the proportion of the images that are predicted to be the $i$-th category while actually belonging to the $j$-th class is computed. Therefore, the confusion matrix can directly visualize the performance of each category and through it we can easily get which classifiers are getting it right and what types of errors they are making. In this survey we only use OA as evaluation criterion because the confusion matrix will take a lot of space.

\begin{equation}\label{7}
\text{OA}=N_{c} / N_{t}
\end{equation}

\begin{equation}\label{8}
\text{AA}=\frac{1}{c} \sum_{i=1}^{c} A_{i}
\end{equation}

\subsection{Performance Comparison}
In recent years, a variety of scene classification algorithms have been published. Here, 27 deep learning-based scene classification methods are selected for performance comparison on three widely-used benchmark data sets. Among the 27 deep learning methods, 3 of them are autoencoder-based methods, 22 of them are CNN-based methods, and 2 of them are GAN-based methods.

\begin{table*}[!th]
	\centering
    \vspace{\baselineskip}
    \vspace{\baselineskip}
	\caption{Overall accuracy ($\%$) comparison of 15 scene classification methods on the NWPU-RESISC45 data set.}\label{table5}
	\resizebox{0.95\textwidth}{!}{
		\renewcommand\arraystretch{1.6}
		\begin{tabular}{|p{3.8cm}<{\centering}|p{3.8cm}<{\centering}|p{1.2cm}<{\centering}|p{2.5cm}<{\centering}|p{2.0cm}<{\centering}|p{2.0cm}<{\centering}|}	
			\hline
			\multicolumn{2}{|c|}{ \multirow{2}*{Method} }&\multirow{2}*{Year} & \multirow{2}*{Publication} &\multicolumn{2}{c|}{Training ratio} \\
			\cline{5-6}
			\multicolumn{2}{|c|}{} & & &10$\%$ &20$\%$\\
			\hline
			
			\multirow{11}*{CNN-based}
			&BoCF \cite{cheng2017bocf}
			&2017	&IEEE GRSL	&82.65$\pm$0.31	&84.32$\pm$0.17 \\
			
			\cline{2-6}
			&MSCP \cite{he2018remote}
			&2018	&IEEE TGRS	&88.07$\pm$0.18	&90.81$\pm$0.13 \\
			
			\cline{2-6}
			&D-CNNs \cite{cheng2018deep}
			&2018	&IEEE TGRS	&89.22$\pm$0.50	&91.89$\pm$0.22  \\
			
			\cline{2-6}
			&HW-CNNs \cite{liu2018hw}
			&2018	&IEEE TGRS	&-	&94.38$\pm$0.17 \\
			
			\cline{2-6}
			&IORN \cite{wang2018iorn}
			&2018	&IEEE GRSL	&87.83$\pm$0.16	 &91.30$\pm$0.17 \\

			\cline{2-6}
			&ADSSM \cite{zhu2018adaptive}
			&2018	&IEEE TGRS	&91.69$\pm$0.22	&94.29$\pm$0.14   \\
			
			\cline{2-6}
			&SF-CNN \cite{xie2019scale}
			&2019	&IEEE TGRS	&89.89$\pm$0.16	&92.55$\pm$0.14 \\
			
			\cline{2-6}
			&ADFF \cite{zhu2019attention}
			&2019	&Remote Sensing	  &90.58$\pm$0.19	&91.91$\pm$0.23 \\
			
			\cline{2-6}
			&CNN-CapsNet \cite{zhang2019remote}
			&2019	&Remote Sensing	  &89.03$\pm$0.21	&89.03$\pm$0.21 \\
			
            \cline{2-6}
			&SCCov \cite{he2019skip}
			&2019	&IEEE TNNLS &89.30$\pm$0.35	&92.10$\pm$0.25\\

            \cline{2-6}
			&DNE \cite{DNE2019}
			&2019	&IEEE GRSL &-	&96.01\\

			\cline{2-6}
			&Hydra \cite{minetto2019hydra}
			&2019	&IEEE TGRS	&92.44$\pm$0.34	&94.51$\pm$0.21  \\
			
			\cline{2-6}
			&Siamese ResNet50 \cite{liu2019siamese}
			&2019	&IEEE GRSL	&-	&92.28    \\
			\hline
			
			\multirow{2}*{GAN-based}
			&MARTA GANs \cite{lin2017marta}
			&2017	&IEEE GRSL  &68.63$\pm$0.22	&75.03$\pm$0.28	\\
			
			\cline{2-6}
			&Attention GANs \cite{yu2019attention}
			&2019	&IEEE TGRS   &72.21$\pm$0.21	&77.99$\pm$0.19\\
			\hline
	\end{tabular}}
\end{table*}
Tables \ref{table3}, \ref{table4}, \ref{table5} report the classification accuracy comparison of deep learning-based scene classification methods on the UC-Merced data set, the AID data set, and the NWPU-RESISC45 data set, respectively, measured in terms of OA.
\subsection{Discussion}
As can be seen from Tables \ref{table3}, \ref{table4}, \ref{table5}, the performance of remote sensing image scene classification has been successively advanced.  In the early days, deep learning-based scene classification approaches were mainly based on autoencoder, and researchers usually use the UC-Merced data set to evaluate autoencoder-based algorithms. As an early unsupervised deep learning method, the structure of autoencoder was relatively simple, so its feature learning capability was also limited. The accuracies of the autoencoder-based approaches had plateaued on the standard benchmarks.
\\ \indent Fortunately, after 2012, CNNs, a powerful supervised learning method, have proved to be capable of learning abstract features from raw images. Despite their powerful potential, it took some time for CNNs to take off in the remote sensing image scene classification domain, until 2015. A short while later, CNN-based algorithms mainly used CNNs as feature extractors, which outperformed autoencoder-based methods. However, only using CNNs as feature extractors did not make full use of the potential of CNNs. Thanks to the release of two large-scale scene classification benchmarks, namely AID and NWPU-RESISC45 in 2017, fine-tuning off-the-shelf CNNs have shown better generalization ability in the task of scene classification than only using CNNs as feature extractors.
\\ \indent Generally, CNN-based methods require large-scale labeled remote sensing images to train CNNs. To deal with this issue, GANs, a novel self-supervised learning method, was introduced into remote sensing image scene classification. Through adversarial training, GANs can model the distribution of real samples and generate new samples. According to the reported accuracy of scene classification in Tables \ref{table3}, \ref{table4}, \ref{table5}, the development of autoencoder-based methods
have reached a bottleneck, CNNs-based methods still dominate and have some upside potential, the performance of GAN-based methods is relatively low on the three benchmarks, and so there remains much room for further improving the performance of  GAN-based methods.
\\ \indent
Moreover, learning discriminative feature representation is one of the critical driving forces that improve scene classification performance. Fusing multiple features \cite{chaib2017deep,fang2019robust}, designing effective cost functions \cite{minetto2019hydra,liu2018hw}, modifying deep learning models \cite{minetto2019hydra,xie2019scale}, and data augmentation \cite{lin2017marta} are all beneficial for attaining better performance. Meanwhile, with the access to large-scale benchmark data sets, it will become smaller for the gap between the scene classification approaches based on supervised learning and the scene classification approaches relied on unsupervised learning.
\\ \indent The release of publicly available benchmarks, such as the UC-Merced data set, the AID data set and the NWPU-RESISC45 data set, makes it easier to compare scene classification algorithms. From the perspectives of data sets, the UC-Merced data set is relatively simple, and the results on the data set driven by CNNs have reached saturation (above 99$\%$ classification accuracy by using the training ratios of 80$\%$). The AID data set is of moderate difficulty. The classification accuracy on the AID data set can reach about 97$\%$ by using 50$\%$ training samples. For NWPU-RESISC45, some advanced methods based on CNNs have reached about  96$\%$ classification accuracy when the training ratio is fixed at 20$\%$. Up to the present, the NWPU-RESISC45 data set is still challenging compared with the UC-Merced data set and the AID data set.
\\ \indent The performance of CNN-based methods depends very much on the quantity of training data, so developing larger-scale and more challenging remote sensing image scene classification benchmarks can further promote the development of data-driven algorithms.

\section{Future Opportunities}
Scene classification is an important and challenging problem for remote sensing image interpretation. Driven by its wide application, it has aroused extensive research attention. Thanks to the advancement of deep learning techniques and the establishment of large-scale data sets for scene classification, scene classification has been seeing dramatic improvement. In spite of the amazing successes obtained in the past several years, there still exists a giant gap between the current understanding level of machines and human-level performance. Thus, there is still much work that needs to be done in the field of scene classification. By investigating the current scene classification algorithms and the available data sets, this paper discusses several potential future directions for scene classification in remote sensing imagery.

(1) Learning discriminative feature representations. Two key factors that influences the performance of scene classification tasks are intraclass diversity and interclass similarity existing in remote sensing images. To tackle the challenges, some representative methods \cite{minetto2019hydra,cheng2018deep,zheng2019deep} have been introduced over the past few years, such as multi-task learning (e.g., unifying classification and similarity/metric learning) and designing/fusing CNNs. Even though these methods are effective to learn discriminative CNN features, the challenges of higher intraclass variation and smaller interclass separability are still not fully solved. These challenges seriously affect the performance of scene classification. In the future, learning more discriminative feature representations to handle the challenges needs to be addressed by various learning ways.

(2) Learning multi-scale features. In the task of remote sensing image scene classification, the same scene/object class can appear in different scales due to the changes in imaging distance and the intrinsic properties of scenes/objects in size, so how to learn multi-scale features has been a crucial and open problem. Some researches \cite{he2019skip,gao2019res2net,lin2017feature,yu2018deep,liu2018m,cheng2020cross-scale} in multi-scale representations have been done over the past few decades, such as multi-scale training, multi-resolution feature fusion, and changing receptive field. However, these existing methods for learning scale-invariance features are far from the capability of human vision and cannot easily respond to the challenge of large variance of scene/object scale. For example, building deeper CNNs in order to extract high-level features has the side effect that small-sized object information is easily discarded. In the future, designing more robust way to extract multi-scale features, especially for small-sized scenes/objects, would be promising for numerous vision tasks.

(3) Multi-label remote sensing image scene classification. In the past few decades, extensive efforts have been made for the task of single-label image classification. However, in the real world, it is extremely common that multiple ground objects will appear in a remote sensing image because of the bird’s-eye imaging method. Therefore, single-label remote sensing image scene classification does not allow for a deep understanding of the intricate content of remote sensing images. In recent years, research has been conducted on multi-label remote sensing image scene classification \cite{hua2019recurrently,chaudhuri2017multilabel,stivaktakis2019deep,hua2020relation,khan2019graph,zegeye2018novel,cheng2018multi-scale}, but it still faces many challenges that need to be further addressed, such as how to exploit the relationship between different labels, how to learn more generalized discriminative features, and how to build large-scale multi-label remote sensing image scene classification data sets.

(4) Developing larger scale scene classification data sets. An ideal scene classification system would be capable of accurately and efficiently recognizing all scene types in all open world scenes. Recent scene classification methods are still trained with relatively limited data sets, so they are capable of classifying scene categories within the training data sets but blind, in principle, to other scene classes outside the data sets. Therefore, a compelling scene classification system should be able to accurately label a novel scene image with a semantic category. The existing data sets \cite{yang2010bag,cheng2017remote,xia2017aid} contain dozens of scene classes, which are far fewer than those that humans can distinguish. Moreover, a common deep CNN has millions of parameters and it tends to over-fit the tens of thousands of training samples in the training set. Hence, fully training a deep classification model is almost impracticable by using currently available scene classification data sets. A majority of advanced scene classification algorithms mainly rely on fine-tuning already trained CNNs on the target data sets or utilizing pre-trained CNNs as feature extractors. Although the transferring solutions behaves fairly well on the target data sets with limited types and samples, they are not the most optimal solution compared with fully training a deep CNN model because the model trained from scratch is able to extract more specific features that are adaptable to the target domain when training samples is large enough. Considering this, developing a new large scale data set with considerably more scene classes for scene classification is very promising.

(5) Unsupervised learning for scene classification. Currently, the most advanced scene classification algorithms generally use fully supervised models learned from annotated data with semantic categories and have achieved amazing scene classification results. However, such fully supervised learning is extremely expensive and time-consuming to undertake because data annotation must be done manually by researchers with expert knowledge of the area of remote sensing image understanding. When the number of scene classes is huge, data annotation may become very difficult due to the massive amount of diversities and variations in  remote sensing images. Meanwhile, the labeled data is generally full of noise and errors, especially for large-scale data sets, since the diverse knowledge levels of different specialists result in different understandings of the same classes of scene. Fully supervised learning can hardly work well without a large data set with clean labels. As a promising unsupervised learning method, generative adversarial networks have been used for tackling scene classification with data sets that lack annotations \cite{duan2018gan,lin2017marta,yu2019attention}. Consequently, it is valuable to explore unsupervised learning for scene classification.

(6) Compact and efficient scene classification models. During the past few years, another key factor in the outstanding progress in scene classification is the evolution of powerful deep CNNs. In order to achieve high accuracy in classification, the layer number of the CNNs has increased from several layers to hundreds of layers. Most advanced CNN models have millions of parameters and require a massive labeled data set for training and high-performance GPUs, which severely limits the deploying of scene classification algorithms on airborne and satellite-borne embedded systems. In response, some researchers are working to design compact and lightweight scene classification models \cite{chen2018training,zhang2019lightweight}. In this area, there is much work to be done.

(7) Scene classification with limited samples. CNNs have obtained huge successes in the field of scene classification. However, most of those models demand large-scale labelled data and numerous iterations to train their parameter sets. This extremely limits their scalability to novel categories because of the high cost of labeling. Also, this fundamentally confines their applicability to rare scene categories (e.g., missile position, military zones), which are difficult to capture. In contrast, humans are adept at distinguishing scenes with little supervision learning, or none at all, such as few-shot \cite{sung2018learning} or zero-shot learning \cite{ye2017zero}. For instance, children can quickly and accurately recognize scene types ranging from a single image on TV, in a book, or hearing its description. The current best  scene classification approaches are still far from achieving the humans’ ability to classify scene types with a few labelled samples. Exploring few-shot/zero-shot learning approachs for scene classification \cite{koch2015siamese,zhai2019lifelong,li2017zero} still needs to be further developed.

(8) Cross-domain scene classification. Current researches have confirmed that CNNs are powerful tools for the task of scene classification and CNN-based methods have attained remarkable performance. However, the big achievements are based on the fact that training and testing data obey the same distribution. What will happen when train and test sets are from different domains? Can CNN models trained on a source domain show good generalization on another target domain? Generally, the performance will drop significantly because there exists a big gap between the source and target domains on data distribution. In fact, these differences between source and target domains are quite common on remote sensing images because of different imaging platforms (e.g., satellites and unmanned aerial vehicles) or different imaging sensors (optical sensors, infrared sensors, and SAR sensors). In the past few years, some researchers have explored cross-domain scene classification to enhance the generalization of CNN models and reduce the distribution gap between the target and source domains \cite{ammour2018asymmetric,othman2017domain,lu2019multisource,song2019domain}. There is much potential for improving domain adaption-based methods for scene classification, such as mapping the feature representations from target and source domains onto a uniform space while preserving the original data structures, designing additional adaptation layers, and optimizing the loss functions.

\section{Conclusions}
Scene classification of remote sensing images has obtained major improvements through several decades of development. The number of papers on remote sensing image scene classification is breathtaking, especially the literature about deep learning-based methods. By taking into account the rapid rate of progress in scene classification, in this paper, we first discussed the main challenges that the current area of remote sensing image scene classification faces with. Then, we surveyed three kinds of deep learning-based methods in detail and introduced the mainstream scene classification benchmarks. Next, we summarized the performance of deep learning-based methods on three widely used data sets in tabular forms, and also provided the analysis of the results. Finally, we discussed a set of promising opportunities for further research.

{\small
\bibliographystyle{ieeetr}
\bibliography{jstars}
}

\begin{IEEEbiography}[{\includegraphics[width=1in,height=1.25in,clip,keepaspectratio]{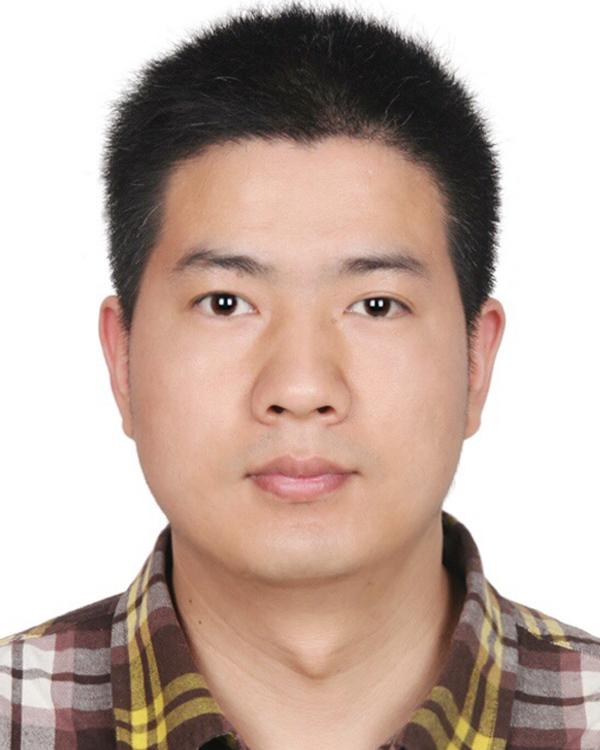}}]{Gong Cheng}
received the B.S. degree from Xidian University, Xi'an, China, in 2007, and the M.S. and Ph.D. degrees from Northwestern Polytechnical University, Xi'an, China, in 2010 and 2013, respectively. He is currently a Professor with Northwestern Polytechnical University, Xi'an, China. His main research interests are computer vision, pattern recognition, and remote sensing image understanding. He is an associate editor of IEEE Geoscience and Remote Sensing Magazine and a guest editor of IEEE Journal of Selected Topics in Applied Earth Observations and Remote Sensing.
\end{IEEEbiography}

\begin{IEEEbiography}[{\includegraphics[width=1in,height=1.25in,clip,keepaspectratio]{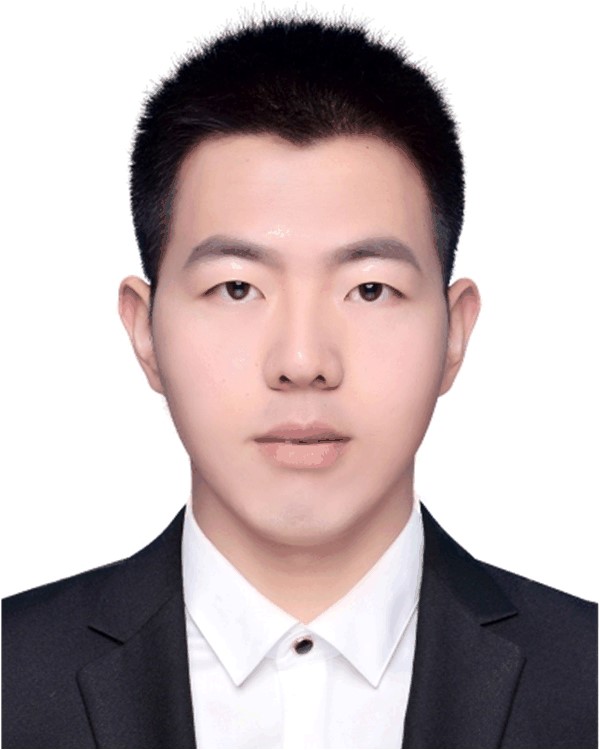}}]{Xingxing Xie}
received the B.S. degree from Inner Mongolia University, Huhhot, China, in 2015, and the M.S. degree from Northwestern Polytechnical University, Xi'an, China, in 2018. Currently, he is pursuing the doctoral degree at Northwestern Polytechnical University. His main research interests are computer vision and pattern recognition.	
\end{IEEEbiography}

\begin{IEEEbiography}[{\includegraphics[width=1in,height=1.25in,clip,keepaspectratio]{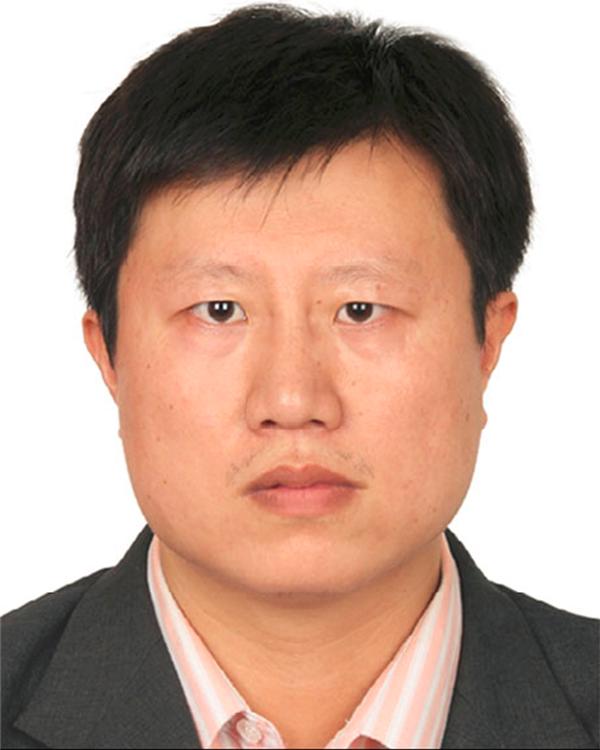}}]{Junwei Han}
received his B.S., M.S., and Ph.D. degrees in pattern recognition and intelligent systems in 1999, 2001, and 2003, respectively, all from Northwestern Polytechnical University, Xi'an, China, where he is currently a professor. He was a research fellow at Nanyang Technological University, The Chinese Univer-sity of Hong Kong, Dublin City University, and the University of Dundee from 2003 to 2010. His research interests include computer vision and brain-imaging analysis. He is an associate editor of IEEE Transactions on Neural Net-works and Learning Systems, IEEE Transac-tions on Circuits and Systems for Video Tech-nology, IEEE Transactions on Human-Machine Systems, Neurocomputing, and Machine Vision and Applications.
\end{IEEEbiography}

\begin{IEEEbiography}[{\includegraphics[width=1in,height=1.25in,clip,keepaspectratio]{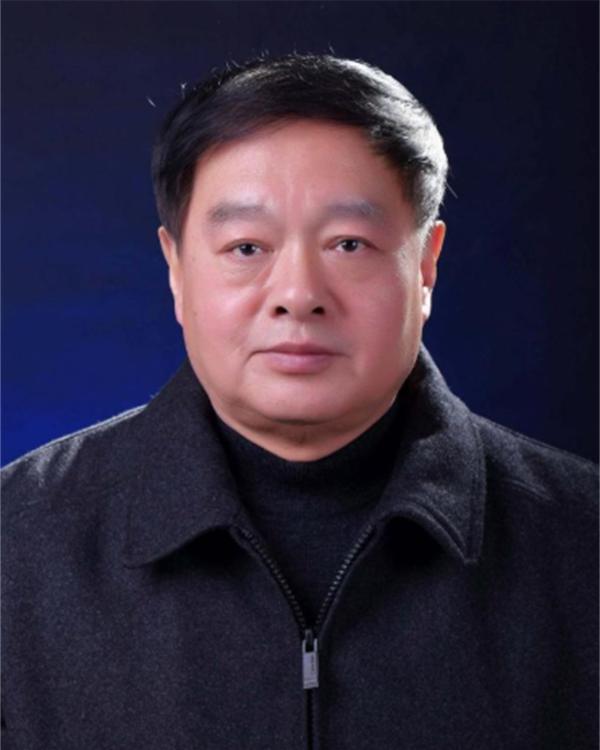}}]{Lei Guo}
received the B.S. and M.S. degrees from Xidian University, Xi'an, China, in 1982 and 1986, respectively, and the Ph.D. degree from Northwestern Polytechnical University, Xi'an, China, in 1993. He is a Professor with the School of Automation, Northwestern Polytechnical University, Xi'an, China. His research interest focuses on image processing.
\end{IEEEbiography}

\begin{IEEEbiography}[{\includegraphics[width=1in,height=1.25in,clip,keepaspectratio]{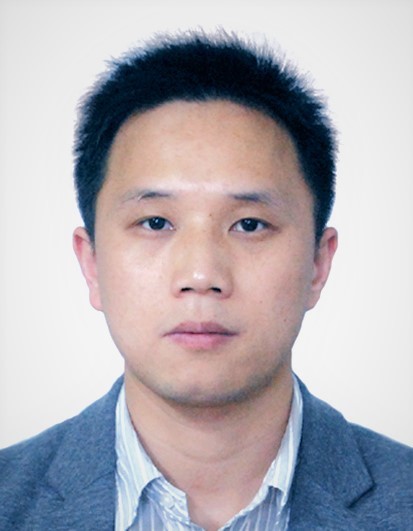}}]{Gui-Song Xia}
(M'10-SM'15) received his Ph.D. degree in image processing and computer vision from CNRS LTCI, T{\'e}l{\'e}com ParisTech, Paris, France, in 2011. From 2011 to 2012, he has been a Post-Doctoral Researcher with the Centre de Recherche en Math{\'e}matiques de la Decision, CNRS, Paris-Dauphine University, Paris, for one and a half years. He is currently working as a full professor in computer vision and photogrammetry at Wuhan University. He has also been working as Visiting Scholar at DMA, {\'E}cole Normale Sup{\'e}rieure (ENS-Paris) for two months in 2018. His current research interests include mathematical modeling of images and videos, structure from motion, perceptual grouping, and remote sensing imaging. He serves on the  Editorial Boards of the journals Pattern Recognition, Signal Processing: Image Communications, and EURASIP Journal on Image \& Video Processing.
\end{IEEEbiography}





\end{document}